\documentclass[sigconf,authorversion,nonacm]{acmart}
\usepackage{framed}
\definecolor{shadecolor}{rgb}{0.94, 0.97, 1.0}
\usepackage{xcolor}

\setcopyright{acmcopyright}
\AtBeginDocument{%
  \providecommand\BibTeX{{%
    \normalfont B\kern-0.5em{\scshape i\kern-0.25em b}\kern-0.8em\TeX}}}

\copyrightyear{2023} 
\acmYear{2023} 
\setcopyright{acmlicensed}\acmConference[KDD'23]{Proceedings of the 29th ACM SIGKDD Conference on Knowledge Discovery and Data Mining}{Aug. 6--10, 2023}{Long Beach, CA, USA}
\acmBooktitle{Proceedings of the 29th ACM SIGKDD Conference on Knowledge Discovery and Data Mining (KDD '23), August 6--10, 2023, Long Beach, CA, USA}
\acmPrice{15.00}
\acmDOI{10.1145/3580305.3599837}
\acmISBN{979-8-4007-0103-0/23/08}

\usepackage{color}
\usepackage{multirow}
\usepackage{autobreak}
\usepackage{caption}
\usepackage[ruled,linesnumbered]{algorithm2e}
\usepackage{subfigure}
\usepackage{makecell}
\SetKwInOut{Parameter}{Parameters}
\usepackage{soul}
\usepackage{balance}

\newcommand{\mymodel}{\textbf{HardSATGEN}}

\usepackage{ulem}

\begin{document}

\title{HardSATGEN: Progressively Generating Your Own Structure and Hardness Aware SAT Formulae}
\title{HardSATGEN: Understanding the Difficulty of Hard SAT Formula Generation and A Strong Structure-Hardness-Aware Baseline}



\author{Yang Li}
\orcid{0000-0002-5249-3471}
\affiliation{%
  \institution{Shanghai Jiao Tong University}
  \city{Shanghai}
  \country{China}
  \postcode{200240}
}
\email{yanglily@sjtu.edu.cn}

\author{Xinyan Chen}
\orcid{0009-0002-7843-5648}
\affiliation{%
  \institution{Shanghai Jiao Tong University}
  \city{Shanghai}
  \country{China}
  \postcode{200240}
}
\email{moss_chen@sjtu.edu.cn}

\author{Wenxuan Guo}
\orcid{0000-0001-6336-3819}
\affiliation{%
  \institution{Shanghai Jiao Tong University}
  \city{Shanghai}
  \country{China}
  \postcode{200240}
}
\email{arya_g@sjtu.edu.cn}

\author{Xijun Li}
\orcid{0000-0002-9013-1180}
\affiliation{%
  \institution{Huawei Noah's Ark Lab}
  \city{Shenzhen}
  \country{China}
}
\email{xijun.li@huawei.com}

\author{Wanqian Luo}
\orcid{0009-0003-2638-1382}
\affiliation{%
  \institution{Huawei Noah's Ark Lab}
  \city{Shenzhen}
  \country{China}
}
\email{luowanqian1@huawei.com}

\author{Junhua Huang}
\orcid{0009-0001-8995-5879}
\affiliation{%
  \institution{Huawei Noah's Ark Lab}
  \city{Shenzhen}
  \country{China}
}
\email{huangjunhua15@huawei.com}

\author{Hui-Ling Zhen}
\orcid{0000-0003-0310-3825}
\affiliation{%
  \institution{Huawei Noah's Ark Lab}
  \city{Hong Kong}
  \country{China}
}
\email{zhenhuiling2@huawei.com}

\author{Mingxuan Yuan}
\orcid{0000-0002-2236-8784}
\affiliation{%
  \institution{Huawei Noah's Ark Lab}
  \city{Hong Kong}
  \country{China}
}
\email{yuan.mingxuan@huawei.com}

\author{Junchi Yan}
\orcid{0000-0001-9639-7679}
\authornote{Junchi Yan is the correspondence author, who is specifically with Department of Computer Science and Engineering, and MoE Key Lab of Artificial Intelligence, SJTU.}
\authornotemark[0]
\affiliation{%
 \institution{Shanghai Jiao Tong University}
  \city{Shanghai}
 \country{China}
  \postcode{200240}
}
\email{yanjunchi@sjtu.edu.cn}


\renewcommand{\shortauthors}{Yang Li et al.}

\begin{abstract}
Industrial SAT formula generation is a critical yet challenging task. Existing SAT generation approaches can hardly simultaneously capture the global structural properties and maintain plausible computational hardness. We first present an in-depth analysis for the limitation of previous learning methods in reproducing the computational hardness of original instances, which may stem from the inherent homogeneity in their adopted split-merge procedure. On top of the observations that industrial formulae exhibit clear community structure and oversplit substructures lead to the difficulty in semantic formation of logical structures, we propose HardSATGEN, which introduces a fine-grained control mechanism to the neural split-merge paradigm for SAT formula generation to better recover the structural and computational properties of the industrial benchmarks. Experiments including evaluations on private and practical corporate testbed show the superiority of HardSATGEN being the only method to successfully augment formulae maintaining similar computational hardness and capturing the global structural properties simultaneously. Compared to the best previous methods, the average performance gains achieve 38.5\% in structural statistics, 88.4\% in computational metrics, and over 140.7\% in the effectiveness of guiding solver tuning by our generated instances. Source code is available at \url{https://github.com/Thinklab-SJTU/HardSATGEN}.
\end{abstract}

    



\maketitle

\begin{snugshade}
\small{
\textcolor{black}{\textit{``CHALLENGE 10: Develop a generator for problem instances that have computational properties that are more similar to real-world instances."\\}
\vspace{2pt}
\textit{\qquad\qquad\qquad\qquad\qquad\qquad\qquad\qquad\qquad--Henry Kautz \& Bart Selman\\Ten Challenges Redux: Recent Progress in Propositional Reasoning and Search\\Principles and Practice of Constraint Programming – CP 2003}}
}
\end{snugshade}

\section{Introduction}
\label{sec:intro}


The \textit{Boolean Satisfiability Problem} (SAT), as the first \textit{Combinatorial Optimization} (CO) problem proven to be NP-complete~\cite{cook1971complexity}, is of central importance in computer science and serves widespread applications in practical scenarios~\cite{hong2010qed,mironov2006applications,ganzinger2004dpll}. Though SAT has an exponential worst-case complexity, the so-known Conflict-Driven Clause Learning (CDCL)~\cite{marques2021conflict} solvers are capable of solving (at least) parts of large-scale industrial SAT benchmarks in a reasonable amount of time. Meanwhile, recent trends have led to a proliferation of machine learning methods solving SAT in a data-driven paradigm~\cite{GuoMIR23}, either as promotive components for heuristics~\cite{selsam2019neurocore,zhangNLocalSATBoostingLocal2020,han2020enhancing} or as end-to-end solutions~\cite{selsam2018learning,GoalNeuralSatICNN22,amizadehLearningSolveCircuitSAT2018}.

Real-world SAT benchmarks may originate from distinct industrial scenarios, exhibiting varying specific structures and solving complexity. Modern SAT solvers have to accommodate complex dependencies to data properties via parameter tuning and strategy selection, while the surging learning-based methods naturally rely on training data to develop their empirical solving policies. However, on the one hand, due to the limited availability of industrial application sources, it can be practically prohibitive to collect adequate quality samples for a specific industrial problem (inherently for a certain distribution or problem structure) at hand. On the other hand, random SAT instances exhibit prominently different properties from industrial benchmarks~\cite{you2019g2sat}, leaving no trivial access to data augmentation. The dilemma heightened the need for efficient industrial formula generation methods~\cite{competition22}.

\cite{yehuda2020s} theoretically shows that generating solid instances for a computationally hard problem is also highly complex and no easier than solving it. Posted as one of the ten critical challenges in propositional reasoning and search~\cite{kautz1997ten,kautz2003ten}, studies on industrial SAT benchmark generation remain limited in number for its difficulty. Previous methods still fall short of reproducing the same (or even similar) properties of original benchmarks in two folds, i.e., mimicking the overall graph structure and achieving similarity in computational hardness. There are primarily two lines of work toward SAT formula generation. Hand-crafted formula generation methods~\cite{giraldez2017locality,giraldez2015modularity} first identify one or two structural metrics and then specify an algorithm capable of controlling these metrics to match the industrial benchmarks. These models fail to unravel specific data characteristics adaptively, merely matching partial structural metrics through manual control. To reveal the underlying overall structural prior from specific applications, a plausible tactic is to employ deep neural nets to automatically learn the structures from specific data~\cite{wu2019learning,you2019g2sat,garzon2022performance}. Learning-based methods naturally accommodate the capture of global graph structures through data-driven training. However, they generally fail to match the hardness (e.g., computational costs) with the specific industrial benchmarks. E.g., the model trained with benchmarks maintaining a solving time of several hundred seconds typically generates formulae that can be solved in seconds. The failure of the computational hardness recovery makes the generation somewhat unreliable for practical applications, such as parameter tuning for SAT solvers.\looseness=-1


Existing learning-based generation methods represent SAT formulae as graphs, e.g., literal-clause graph (LCG), recasting the original problem as a graph generation task~\cite{wu2019learning}. In particular, a node split-merge framework has been commonly adopted as a standard pipeline for SAT formula generation~\cite{you2019g2sat,garzon2022performance} which enjoys flexibility and interpretability. The framework can be efficiently trained via supervised learning whereby the binary labels are collected from the split stage, indicating which pairs shall be merged in the merge phase. While the test refers to merging nodes on the split graph templates to produce new formulae as the generation.

In this paper, we observe and argue that the failure of existing neural methods to reproduce the hardness can specifically result from the inability to imitate and construct rigorous logical substructures. By a careful revisit to the split-merge framework, we make two observations of industrial SAT benchmarks on the structure and computational hardness, respectively: 
    

\textbf{1) Industrial formulae exhibit community structure correlated with hardness:} Industrial SAT formulae characteristically exhibit a clear community structure (i.e. high modularity~\cite{newman2004finding}) in contrast to randomly generated ones as concluded in ~\cite{giraldez2015modularity}. In other words, it is implied that for graph representation of formulae, the nodes can be easily grouped into sets such that each set of nodes is densely connected internally and sparsely connected externally. Also, it is reported that the community structure (modularity) is extensively correlated with the runtime of CDCL-based solvers~\cite{giraldez2015modularity}, which reflects the hardness of the problem.

    
\textbf{2) Oversplit substructures lead to the difficulty of semantic formation of structures:} Unsatisfiable cores, defined as a subset of clauses whose conjunction is still unsatisfiable, have shown prominent effects on the computational hardness of unsatisfiable instances~\cite{lynce2004computing}. The cores possess rigid logical structures, and it is more feasible to form such structures from substructures with certain semantics. While existing learning-based works oversplit the node into very preliminary pieces, making merging them into meaningful structures difficult, thus leading to hardness collapse.

In light of the above characteristics, we propose a multi-stage substructure merging pipeline for SAT formula generation starting from templates produced by unsatisfiable formulae entitled {\mymodel} on top of the  node split-merge framework~\cite{you2019g2sat} with the following devised key steps: \textbf{1)} Controllably scrambling the core by shuffling the variable, clauses and phases of literals; \textbf{2)} Developing connections via clause merging within communities; \textbf{3)} Developing connections in the between of the communities.



The scrambling operation ensures the consistency of the strict logical structure, while randomly shuffling variables, clauses and phases of literals to pose impacts on the computational performance of SAT solvers. To enable community-controlled generation, the prior community division is introduced through community detection on the VIG representation of the formulae. Communities are represented as an additional node set to the bipartite graph, where each node connects its own member variables in VIG. Set operations on the second-order neighbors of community nodes enable the control of community-based phase separation while embedding community knowledge into graphs equips the model with community awareness. A delicately designed post-processing is enforced to ensure the dominance of the scrambled core on the computational performance and can also serve to alter the phase from unsatisfiable to satisfiable. \textbf{The highlights are as follows:}

\textbf{1) Understanding the difficulty of hardness reproducing for SAT formula generation:} We present a concrete analysis for the hardness degradation issue for existing learning-based methods: the inherent homogeneity in the split-and-merge procedure for sample generation. Specifically, the graph is over-split into very preliminary starting substructures with limited diversity, of which a few are prone to be merged by the  learned aggregation policy with high probability, also leading to limited diversity of the newly merged substructures. The hardness is especially collapsed when it is solved with current mainstream CDCL solvers, as they perform logic inference in a backtracking way that fits well with the generated problem's homogeneous structure.
\textbf{2) Novel fine-grained multi-stage merging pipeline aware of both structure and hardness:} We propose {\mymodel} which introduces a fine-grained control mechanism over the community structure and the cores to the split-merge paradigm for SAT formula generation. Specifically, we propose a multi-stage generation pipeline involving core scrambling, in-community connection, and cross-community connection to guide the iterative split-merge procedure for better similarity in structure and hardness.
\textbf{3) Strong empirical results on public and private corporate datasets:} Extensive experiments on public benchmarks and industrial testbed from a worldwide corporate demonstrate the significant superiority of {\mymodel} regarding structure and hardness mimicking, compared to state-of-the-art learning-based and hand-crafted methods for SAT formula generation. To our best knowledge, it is the first work  achieving similar hardness for industrial instances.\looseness=-1





\section{Preliminaries and Related Works}\label{sec:related}
\subsection{SAT Problem and Its Graph Representation}
SAT problem is to determine whether there exists an assignment to satisfy a specific Boolean formula. The formulae consist of Boolean variables connected by logic operators, including conjunction ($\wedge$), disjunction ($\vee$), and negation ($\neg$). Boolean formulae are generally represented in Conjunctive Normal Form (CNF) as a conjunction of clauses, where a clause is a disjunction of literals and a literal denotes either a variable or its negation. SAT also plays an important role in data mining applications, e.g. Maximal Frequent Subgraph Mining~\cite{liu2022satmargin}, Explainable AI~\cite{boumazouza2021asteryx}, Sequence Mining~\cite{jabbour2013boolean}.
Graph representations play a significant role in SAT formulae analysis, which basically has four forms~\cite{biere2009handbook}: literal-clause graph (LCG), literal-incidence graph (LIG), variable-clause graph (VCG), and variable-incidence graph (VIG). LCG is a bipartite graph composed of a literal node set and a clause node set, and an edge therein indicates the occurrence of the literal in the clause. A LIG consists of literals as the nodes, where an edge indicates the co-occurrence of two literals in one clause. VCG and VIG can be derived from LCG and LIG by merging the literals with their negations. This paper's diagrams are generally based on VCG for its simplicity and comprehensibility.\looseness=-1

\subsection{Learning for SAT/CO Solving}
Recently, learning-based methods have shown effectiveness in exploring heuristics automatically from extensive data, advancing the CO field~\cite{GuoMIR23,geng2023vnep,WangICML23}. There are basically two routes of the learning-based methods for SAT: 1) \cite{selsam2018learning,GoalNeuralSatICNN22,amizadehLearningSolveCircuitSAT2018} building end-to-end neural SAT solvers for relatively small-scale instances through graph neural networks; 2) \cite{selsam2019neurocore,zhangNLocalSATBoostingLocal2020,han2020enhancing,li2022kissat} modifying existing solvers with learning-aided components to improve the performance of SOTA solvers.\looseness=-1

\textbf{Remark.} Despite the above progress, so far learning-based solvers can yet only process very small random instances compared to industry benchmarks. In this paper, we mainly focus on the industrial instance generation part with established metrics to evaluate the effectiveness of our generator and follow the current standard protocol~\cite{you2019g2sat,Cowen-Rivers2022-HEBO} by hyperparameter tuning to testify its value for improving  well-established non-learning SAT solvers~\cite{audemard2009glucose,fleury2020cadical}. 

\begin{figure}[tb!]
    \centering
    \includegraphics[width=0.85\linewidth]{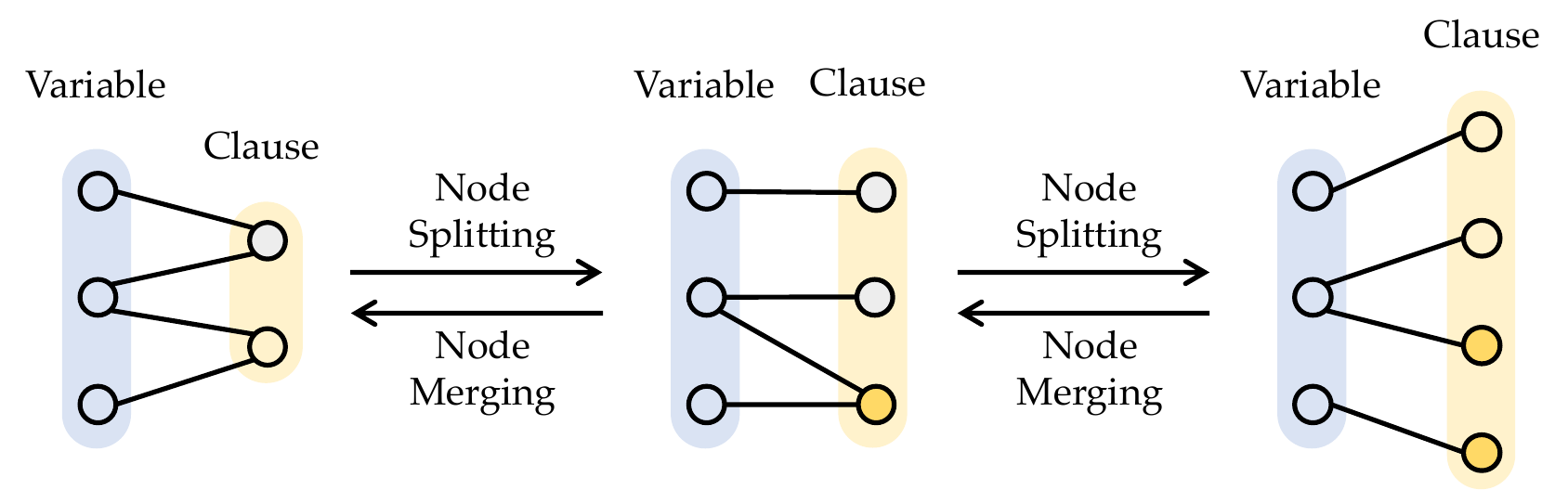}
    \vspace{-8pt}
    \caption{Node splitting and merging operations on graphs.}
    \label{fig:split-merge}
    \vspace{-5pt}
\end{figure}

\subsection{SAT Generators and the Node Split-Merge Framework for ML Generators}

\textbf{Miscellaneous methods: learning-free and learning-based.} To resolve the data bottleneck, two lines of works dominate the community. Hand-crafted formulae generators study the characteristics of real-world benchmarks and devise targeted algorithms for matching one or two structural metrics of interest. SF~\cite{sf09ijcai} focuses on the scale-free structure, i.e., the number of variable occurrences generally follows a power-law distribution. CA~\cite{giraldez2015modularity} concentrates on the community structure and matches the modularity metric. PS~\cite{giraldez2017locality} attempts to achieve an approximation of both properties through a unified pipeline. In general, handcrafted generators are limited to the heuristics and cannot fully leverage the data at hand. In particular, it is difficult for them to imitate and achieve similar structure and hardness for arbitrary input samples. Recently machine learning has been leveraged to advance the generation of SAT formulae beyond the expert rules~\cite{GuoMIR23}. The seminal work SATGEN~\cite{wu2019learning} adopts pre-existing graph generative models to generate LIG representation of SAT formulae. However, LIG graphs cannot directly be translated to formulae since LIG representation is not a bijective mapping of formulae. 

\textbf{One-stage Node Split-Merge Framework based methods.} The seminal G2SAT~\cite{you2019g2sat} resolves SATGEN's issue by developing a one-stage node split-merge framework for bipartite graphs to generate LCG representations of formulae. This framework has also been widely adopted as the foundation of many generation methods including GCN2S, EGNN2S, and ECC2S proposed in~\cite{garzon2022performance}, whose differences lie in the specific design of graph representation and graph neural nets (GNNs).

As shown in Fig.~\ref{fig:split-merge}, the framework~\cite{you2019g2sat} is based on the observation that bipartite graphs can be split into a set of 2-layer depth trees by splitting a node and passing its partial connections to a new node. Thus, generating a bipartite graph can be boiled down to sequentially merging trees. Based on the trees obtained by splitting real SAT formulae, SAT formulae can be generated by sequentially merging clause nodes that integrate two clause nodes' connections to form a new clause node, and a GNN is trained to decide which node pair shall be merged at each step.


Unfortunately, the works under this framework are still not yet able to produce hard instances as industry benchmarks for their inherent homogeneity as will be discussed in detail in Sec.~\ref{sec:whats}, e.g., G2SAT merely verifies its solver performance mimicking ability through reproducing different solvers' ranking of runtime, and in fact the instances generated by G2SAT are rather easier to solve, e.g., typically seconds for modern CDCL solvers. Meanwhile, there still leaves room for boosting the performance in structural similarity. In this paper, we resort to a multi-stage framework,  by proposing fine-grained controls over the community structure and the cores.

\section{Existing One-stage Split-Merge's Limitation in Reproducing Hardness}
\label{sec:whats}

Although the one-stage node split-merge framework~\cite{you2019g2sat} fits well with the bipartite graph generation task and has proven its effectiveness in structure mimicking, in SAT formula generation, the generated instances are trivially easy for modern SAT solvers. In this section, we analyze and argue that the previously mentioned hardness degradation issue can be derived from the inherent homogeneity in the split-merge procedure for sample generation.


Specifically, in the template preparation phase, each clause node continues to be split, the degree of which is consequently decreasing till one during the splitting to produce the templates for generation. This process ends up with clause nodes that merely connect with a single variable node, and for each time one variable appears in the original formula there will be a clause node connected to it (see Fig.~\ref{fig:split-merge}). Thus the templates can eventually maintain a large number of repetitive clause nodes connecting to semantically the same variable. In other words, the graph is over-split into very preliminary starting substructures with limited diversity.

On the other hand, the node merging process is guided by the learned aggregation policy where the node pairs that are prone to be merged maintain a relatively high predicted probability. With limited control, the unified policy on preliminary  substructures can lead to common patterns of merged substructures and finally organize a tree-like logic structure of the generated formula. The generated homogeneous structure fits well with the typical solving paradigm of CDCL solvers which perform logic inference in a backtracking manner, leading to the hardness collapse.


\begin{figure}[tb!]
    \centering
	\includegraphics[width=0.9\linewidth]{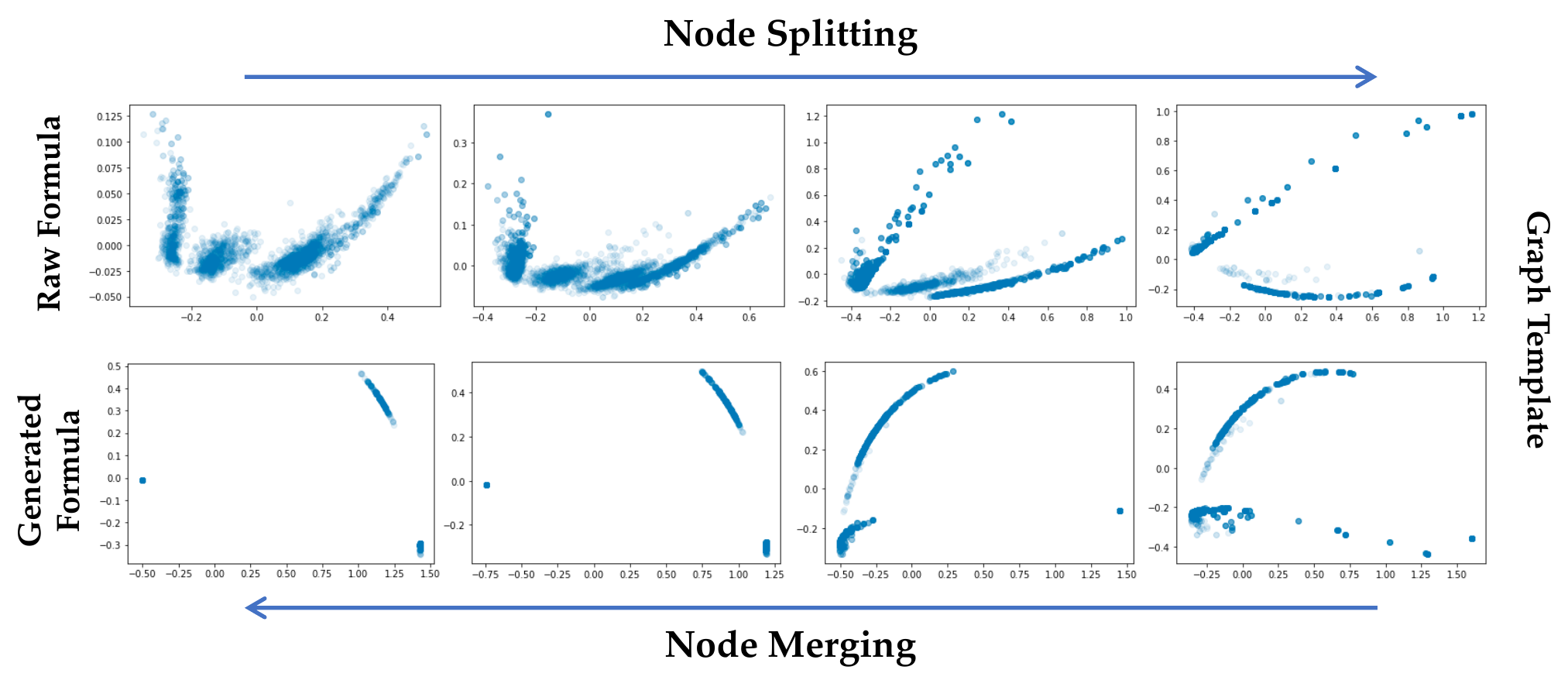}
    \vspace{-10pt}
    \caption{Visualization of clause feature distributions during node split-merge process. The features are extracted by the learned GNN and the dimension reduction is performed by PCA~\cite{pearson1901liii}. The clause features tend to collapse to certain patterns since the clauses presumably exhibit similar structures.}
     \label{fig:feat-dist}
     \vspace{-5pt}
\end{figure}

To support the claim that one-stage split-merge generation eventually maintains clauses with shared substructures of variable combinations, we visualize the shift of the clause feature distribution during the overall pipeline. Fig.~\ref{fig:feat-dist} presents the visualization results. To enable an intuitive analysis of the distribution, we use PCA for dimensionality reduction to 2D, and the x- and y-axes correspond to the first and second principal components.  As seen, the real-world formula's clause feature distribution is relatively scattered with an underlying structural organization. During the splitting, the clause features tend to aggregate into sparse and discrete points which may specifically correspond to the repetitive preliminary starting substructures in the graph template. Through the node merging process, the generated graph fails to reproduce the feature distribution and the clauses promptly tend to collapse to specific feature patterns, since they may extensively share common substructures.

There can be two reasonable approaches to resolving the dilemma. On the one hand, ceasing the splitting process when it reaches some rigorous structure can preserve more key structures and depart from the tree-like logical organization, while retaining rigorous logical structures can presumably preserve computational hardness. On the other hand, the similarly merged substructures produced by the aggregation policy can be moderated by posing more controls on the node merging process, e.g., node pair proposals, which confine the aggregation to limited available options and reduce the potential consecutive merging of similar substructures.

\section{{\mymodel}: Structure and Hardness Aware Generation Pipeline}
\label{sec:method}

This section presents our structure and hardness aware generation pipeline. Sec.~\ref{sec:build} introduces the proposed bipartite graph representations that embed prior knowledge about the community and core, then built upon the representations, Sec.~\ref{sec:split} and ~\ref{sec:infer}  demonstrate the preparation phase for training data and graph templates and the main-body generation pipeline. Finally, Sec.~\ref{sec:train} shows the training scheme for the GNN modules utilized in instance generation.\looseness=-1



\begin{figure}[tb!]
    \centering
	\includegraphics[width=0.5\linewidth]{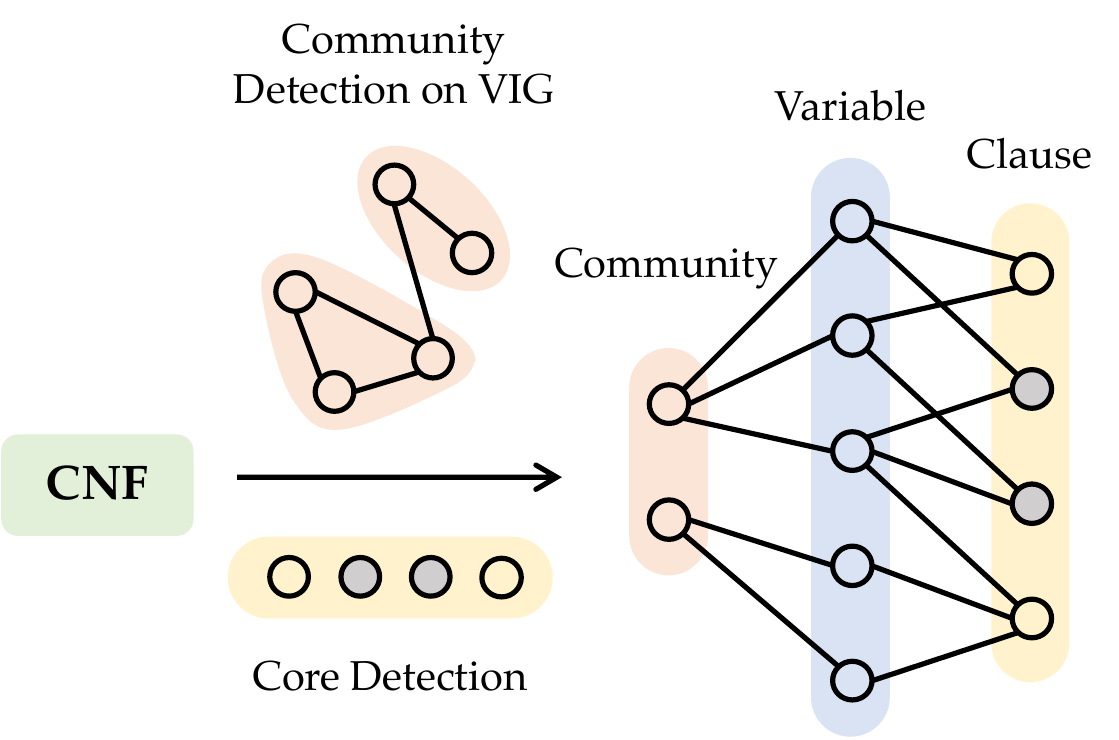}
    \vspace{-5pt}
    \caption{Visualization of the graph building process.}
     \label{fig:graph_building}
     \vspace{-5pt}
\end{figure}

\subsection{Graph Building: Community and Core Embedded Bipartite Graph Representations} \label{sec:build}

Recall that LCG/VCG graph representations model a bipartite graph with node sets of literals/variables and clauses, the edges indicating the occurrence of the literal/variable in the clause. For unsatisfiable formulae, we identify the community structure and the core as two critical components determining the structural and computational properties of the formulae. Thus, in this paper, we introduce additional prior knowledge about community division on VIG and the detected core to the graph representations and thus producing community and core embedded LCG/VCG graph representations. 

For an input CNF, we first apply community detection on VIG representation of the formula by Clauset-Newman-Moore greedy modularity maximization~\cite{clauset2004finding}, then for each detected community, we assign a community node and connect it to its affiliated variable/literal nodes. The resulting graph consists of three node sets representing communities, literals/variables, and clauses, where edges only exist in the between of communities and literals/variables or in the between of literals/variables and clauses. Moreover, we conduct core detection on top of the raw formula by the specialized tool DRAT-trim~\cite{wetzler2014drat}, which identifies a subset of the self-unsatisfiable clauses. Fig.~\ref{fig:graph_building} shows the graph building process.

Taking variable-clause graph as an example, the graph representation can be formulated as $G=(\mathcal{V}, \mathcal{E})$ where $\mathcal{V}=\mathcal{V}_{cmty}\cup\mathcal{V}_{var}\cup\mathcal{V}_{cl}$ and $\mathcal{E}\subseteq(\mathcal{V}_{cmty}\times\mathcal{V}_{var})\cup(\mathcal{V}_{var}\times\mathcal{V}_{cl})$. Here $\mathcal{V}_{cmty}$ denotes the community nodes, $\mathcal{V}_{var}$ denotes the variable nodes, and $\mathcal{V}_{cl}$ denotes the clause nodes. $\mathcal{V}_{cl}^{\prime}\subseteq\mathcal{V}_{cl}$ stands for the clauses in the core. The edges can only connect community nodes and variable nodes for the affiliation relation, or connect variable nodes and clause nodes for the occurrence of the variable in the clause. 

\subsection{Node Splitting: Preparation for Training Data and Graph Templates}\label{sec:split}

The node splitting phase serves two major purposes, i.e., collecting training data and producing graph templates. In the methodology, GNN serves to predict the probability of the node pair being merged, thus node pairs corresponding to a specific graph shall be provided as the training data, which can be obtained via the splitting process. The generation pipeline is performed on the graph templates, which can be produced as the final split results of the formulae.
\begin{figure}[tb!]
    \centering
	\includegraphics[width=0.75\linewidth]{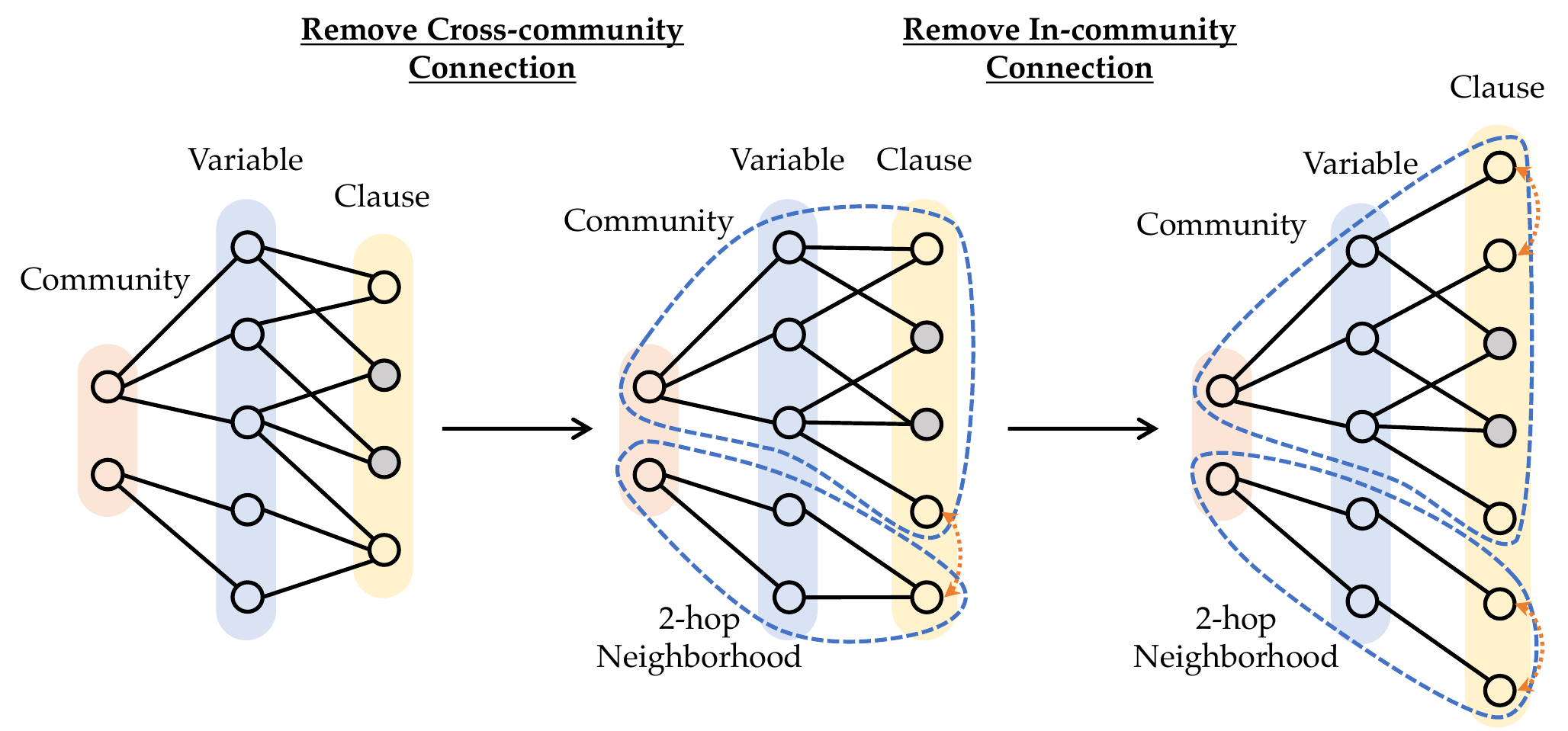}
    \vspace{-10pt}
    \caption{Visualization of node splitting process.}
     \label{fig:splitting}
     \vspace{-5pt}
\end{figure}

Based on the discussion in Sec.~\ref{sec:whats}, beyond the one-stage node split-merge framework, we retain the core as the key logical substructure and pose controls on the community structure. Fig.~\ref{fig:splitting} shows the splitting process. The core in the graph representation is the sub-graph composed of $\mathcal{V}_{cl}^{\prime}$, its 1-hop neighbors $\mathcal{V}_{var}^{\prime}$, and the related edges $\mathcal{E}^{\prime}\subseteq\mathcal{V}_{cl}^{\prime}\times\mathcal{V}_{var}^{\prime}$. The splitting operations are performed on $\mathcal{V}_{cl}-\mathcal{V}_{cl}^\prime$ such that the core can be retained. For controlling the community structure, we separate the splitting phase into two stages removing cross-community and in-community connections respectively, producing the corresponding two-stage training data and allowing the community-phased generation.


In the first stage, for a randomly picked clause node $v$, the model will check its 2-hop neighbors $\mathcal{N}_{v}^{(2)}=\{c_1,\cdots,c_k\}$ to identify the number of communities the adjacent variables involve. If the adjacent variables are involved by more than one communities, it then picks one community, removing the edges between $v$ and the adjacent variable(s) belonging to this community, e.g. variables in $\mathcal{N}_{v}^{(1)}\cap\mathcal{N}_{c_1}^{(1)}$, and assigns those edges to a new node. This process continues till $\forall i,j, \mathcal{N}_{c_i}^{(2)}\cap \mathcal{N}_{c_j}^{(2)}=\emptyset$. In the second stage, the remaining clauses in $\mathcal{V}_{cl}-\mathcal{V}_{cl}^\prime$ will be split indiscriminately, and the graph templates are then produced. The above process produces positive training samples $(u^+,v^+, G)$ for in-community  and cross-community connection respectively. For negative samples, at each splitting, besides the split positive node pair $(u^+,v^+)$ in $G$, we also select $v^-$ to form a negative node pair $(u^+, v^-)$. The selection of $v^-$ shares the same affiliation of $v^+$ regarding the community, e.g., $v^-\in \mathcal{N}_{c}^{(2)}\cap \mathcal{V}_{cl}$ where $c$ satisfies $u^+, v^+\in \mathcal{N}_{c}^{(2)}\cap \mathcal{V}_{cl}$ in the in-community stage, and $v^-\in \mathcal{V}_{cl} - \mathcal{N}_{c}^{(2)}$ where $c$ satisfies $u^+, v^+\in \mathcal{N}_{c}^{(2)}\cap \mathcal{V}_{cl}$ in the in-community stage. The data tuples $(u^+,v^+,v^-,G)$ eventually form the training datasets.

\subsection{Model Inference: General Multi-Stage Instance Generation Pipeline}\label{sec:infer}
\begin{figure*}
    \centering
    \includegraphics[width=0.90\linewidth]{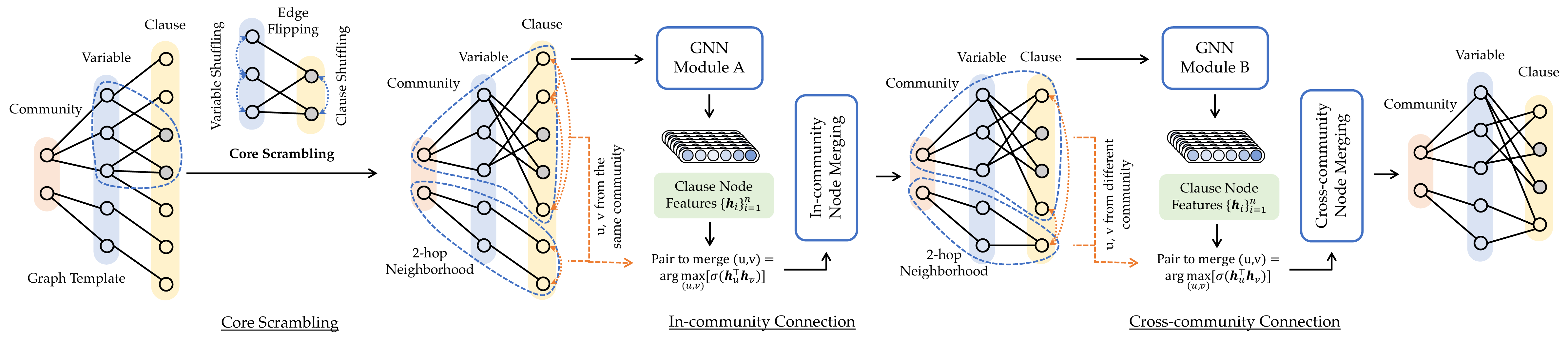}
    \vspace{-12pt}
    \caption{The generation (i.e. the inference phase) involves conducting core scrambling, in-/cross-community connection on the graph templates, which are split from benchmark instances. The connections are performed by sequentially merging the predicted node pair, where the node pair proposals are controlled by set operations on 2-hop neighborhoods of the community.}
    \vspace{-5pt}
    \label{fig:inference}
\end{figure*}

This subsection specifically introduces {\mymodel}'s multi-stage generation pipeline involving core scrambling, in-community connection and cross-community connection, as shown in Fig.~\ref{fig:inference}. We denote the template which is split from the real formulae as $G_0$.



\subsubsection{Core Scrambling with Tie-in Post-processing} \label{subsec:core}

The scrambling process is enforced on the core and produces the resulting graph $G_0^{\prime}$. The core scrambling operation on SAT formulae generally involves the random permutation of variables and clauses, together with random flipping of the literals. While in this paper, we transform the scrambling operation on graph representations, which is composed of reordering the set of variable nodes, reordering the set of clause nodes, and flipping the bipartite graph edge attributes (for weighted edges) or swapping the connected sets (for unweighted edges).

The scrambling operation can be interpreted as the randomization of the core  without directly disrupting the original rigorous logical structure. It was observed in the SAT Competition~\cite{le2004essentials}, which first conducted random scrambling of formulae in verification, that such scrambling might result in totally different runtimes of the same solver, entitled Lisa Syndrome~\cite{le2004essentials}. Thus, conducting core scrambling can pose impacts on the computational properties of the formulae while retaining the rigorous logical substructure, fitting well with the industrial formulae augmentation task.

\begin{figure}[tb!]
    \centering
    \includegraphics[width=0.72\linewidth]{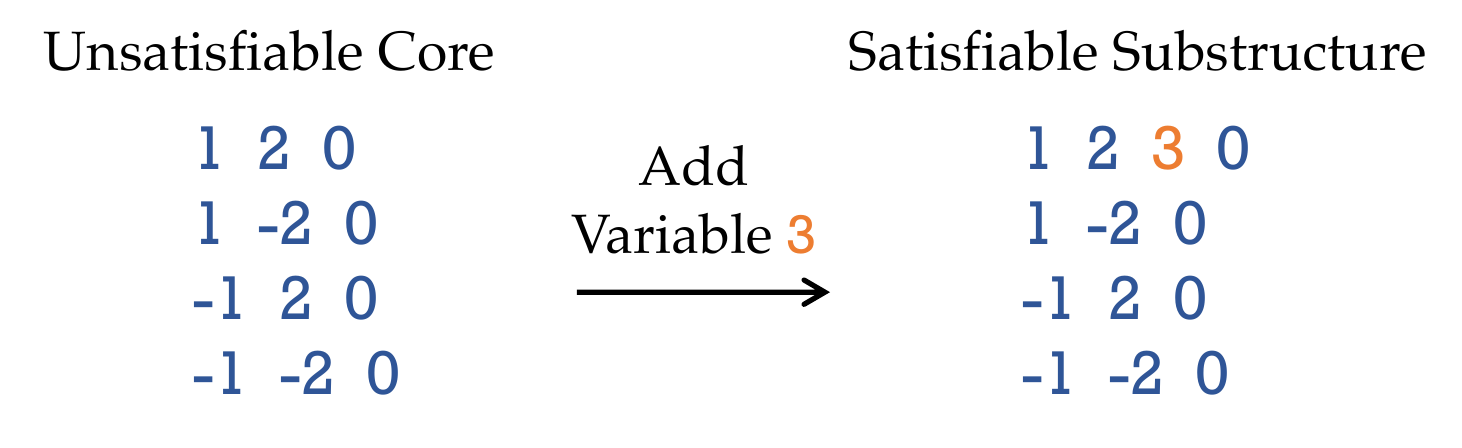}
    \vspace{-10pt}
    \caption{We devise a novel post-processing step that iteratively transforms unexpected small cores to satisfiable substructures by introducing a new variable to one of its clauses.}
    \vspace{-5pt}
     \label{fig:postprocess}
\end{figure}

We deploy a tunable operation for randomizing the cores which is capable of specifying the amount of randomization explicitly. Former works have shown the extent of randomization can affect computational properties of the formulae, where over-randomization may destroy the formula's structure and make the formula harder~\cite{biere2019effect}. The tunability on the extent of randomization can implicitly serve to pose controls over the similarity of computational properties to the original formulae. Specifically, we enforce a real value tuple $(P_1, P_2, P_3)$ where $0\leq P_1, P_2, P_3\leq 1$ to control the extent of randomization for variable permutation, clause permutation, and flipping of the literals, respectively. Taking variable permutation as an example, while traversing the variables, the model randomly exchanges the variable with a distinct variable with probability $P_1$ and retains it with probability $1-P_1$. Similar operations are performed on the clauses. For flipping the literals, the model randomly flips the edge attributes (for weighted edges) or swaps the connected sets (for unweighted edges) with probability $P_3$.

To avoid the coupling effect of the subsequently generated edges on the scrambled cores which may  break down the core and thus lead to hardness collapse, we introduce post-processing at the end of the generation, which includes several rounds of core detection. When an unexpected small core is detected, we enforce a new variable to the generated formula and add it to one arbitrary clause that is included by the unexpected core and excluded by the original core, as shown in Fig.~\ref{fig:postprocess}. The processing is performed in raw CNF format, where positive integers represent variables and their negation represents the corresponding negation. The clauses are encoded as a sequence of decimal numbers ending with 0. Through the processing, unnecessary substructures can be iteratively removed. This post-processing can also be enforced on the core to generate satisfiable formulae based on unsatisfiable templates.\looseness=-1


\subsubsection{In-community and cross-community connection.} \label{sec:cmty-connect}
The primary goal of generation is to learn the distribution of graphs $p(G)$. With the scrambled graph $G_0^{\prime}$ as the prior information, we aim to estimate  $p(G|G_0^{\prime})$ via graph networks. We separate this process into two phases as $n_1$-step in-community connection and $n_2$-step cross-community connection. Let $p_1$ and $p_2$ denote the conditional graph distribution of in-community connection and cross-community connection respectively, the generation can be formulated as:
\begin{small}
\begin{equation}
\begin{aligned}
    p(G|G_0^{\prime}) &= p(G_{n_1+n_2}|G_0^{\prime})
    = p(G_{n_1}|G_0^{\prime})p(G_{n_1+n_2}|G_{n_1})\\
    &=\prod_{i=1}^{n_1} p_1(G_i|G_0^{\prime},\cdots, G_{i-1})\prod_{i=n_1+1}^{n_1+n_2} p_2(G_i|G_{n_1}, \cdots, G_{i-1})
\end{aligned}
\end{equation}
\end{small}
Here $G_i$ denotes the intermediate graph at step $i$. Yet we make a basic assumption that the ordering of the intermediate graphs does not affect the resulting graphs such that the  \textit{Markov} property is satisfied. Then we have:
\begin{small}
\begin{align}
    p(G|G_0^{\prime}) 
    &=p_1(G_1|G_0^{\prime})\prod_{i=2}^{n_1} p_1(G_i|G_{i-1})\prod_{i=n_1+1}^{n_1+n_2} p_2(G_i|G_{i-1})
\end{align}
\end{small}
Following the node split-merge paradigm, we model $p_{\cdot}(G_i|G_{i-1})$ as the node merging operation on $G_{i-1}$, e.g., merging nodes $u$ and $v$ on $G_{i-1}$, which is denoted as $\operatorname{NodeMerge}(G_{i-1},u,v)$. Then respectively for in-community and cross-community connection:
\begin{footnotesize}
\begin{equation}
\begin{aligned}
    p_1(G_i|G_{i-1})&=p_1(\operatorname{NodeMerge}(G_{i-1},u,v)|G_{i-1})\\
    &=\operatorname{Multinomial}(p_1^{\prime}(u,v)|C(u)=C(v),\forall u,v \in \mathcal{E}_{cl}-\mathcal{E}^{\prime}_{cl})
\end{aligned}
\end{equation}
\begin{equation}
\begin{aligned}
    p_2(G_i|G_{i-1})&=p_2(\operatorname{NodeMerge}(G_{i-1},u,v)|G_{i-1})\\&=\operatorname{Multinomial}(p_2^{\prime}(u,v)|C(u)\neq C(v),\forall u,v \in \mathcal{E}_{cl}-\mathcal{E}^{\prime}_{cl})
\end{aligned}
\end{equation}
\end{footnotesize}
Here $C(\cdot)$ denotes the community that the node belongs to and $p_{\cdot}^{\prime}(u,v)$ denotes the probability of $(u,v)$ to be merged. We model $p_{\cdot}^{\prime}(u,v)$ by the sigmoid of the inner product of node features $\sigma(\mathbf{h}_u^\top\mathbf{h}_v)$. The features are extracted by GNN modules and two GNNs are trained respectively to learn the distribution for in-community connection $p_1^{\prime}$ and cross-community connection $p_2^{\prime}$. 

For a node pair candidate, we control it by the set operations on the 2-hop neighbors $\mathcal{N}_{c_i}^{(2)}$ of the community nodes $c_i$. For in-community connection, node pairs shall be sampled from $\bigcup_i(\mathcal{N}_{c_i}^{(2)}\times \mathcal{N}_{c_i}^{(2)})$, i.e., pairs within the same community. For cross-community connection, node pairs shall be sampled from $(\mathcal{E}_{cl}-\mathcal{E}^{\prime}_{cl})\times (\mathcal{E}_{cl}-\mathcal{E}^{\prime}_{cl}) - \bigcup_i(\mathcal{N}_{c_i}^{(2)}\times \mathcal{N}_{c_i}^{(2)})$, i.e., pairs connects different communities.

The number of steps for in-community and cross-community connection denoted as $n_1$ and $n_2$ are by default set to be equal to the step number of in-community splitting and cross-community splitting denoted as $m_1$ and $m_2$. However, the phase transition point is potentially tunable through the parameter $n_1$ and $n_2$. This enables the model to control the graph structural metrics (especially the modularity reflecting the community structure) of the generated formulae within a certain margin. In implementation, we introduce $\alpha$ to explicitly control the structural properties, where $n_1$ and $n_2$ are derived by $n_1 = m_1 + (1-\alpha)m_2$ and $n_2 = \alpha m_2$. By default, $\alpha=1$. A larger $\alpha$ produces graphs with smaller modularity and vice versa.

\subsection{Model Training: Learn to Predict Merging Pairs with Two-stage Control}\label{sec:train}

\begin{figure}[tb!]
    \centering
	\includegraphics[width=0.8\linewidth]{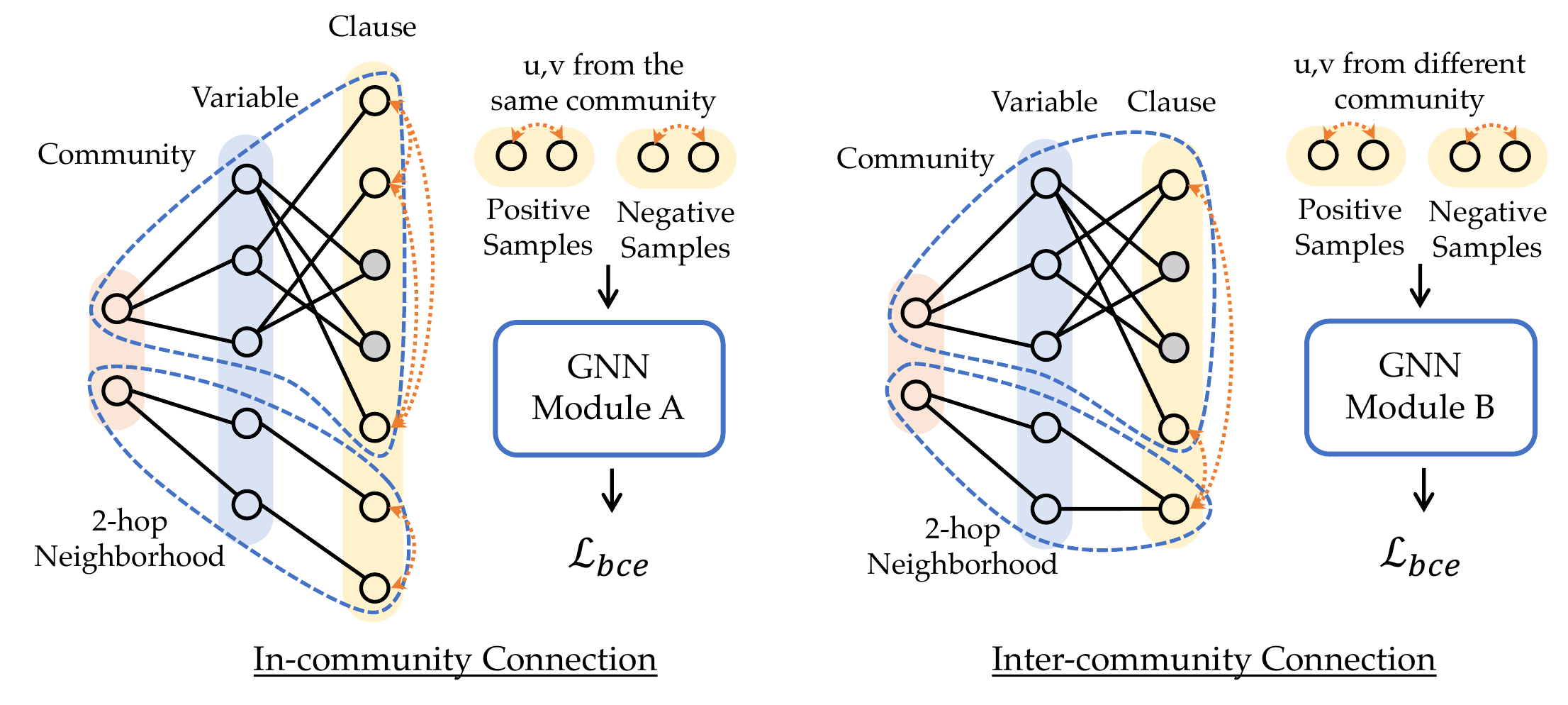}
    \vspace{-10pt}
    \caption{Visualization of the GNN training process.}
     \label{fig:training}
     \vspace{-5pt}
\end{figure}

Fig.~\ref{fig:training} shows the training pipeline, which consists of two GNN modules to learn the structural features and predict the probability for node pairs to be merged. Following \cite{you2019g2sat}, we formulate the learning task as a binary classification to distinguish whether the specific node pair should be merged or not, and we further introduce a two-stage control and train two GNN modules to capture the features for in-community  and cross-community connection respectively.

\subsubsection{Graph Network Design} Since our  design is orthogonal to the graph network backbone selection, we verify the generation performance of {\mymodel} on two mainstream network backbones for node embedding  including GCN~\cite{kipf2016semi} and GraphSAGE~\cite{hamilton2017inductive}. Specifically, the $l$-th layer of GCN is formulated as:
\begin{small}
\begin{equation}
\begin{aligned}
\textbf{H}^{(l+1)}=ReLU(\tilde{D}^{-\frac{1}{2}}\tilde{A}\tilde{D}^{-\frac{1}{2}}\textbf{H}^{(l)}\textbf{W}^{(l)})
\end{aligned}
\end{equation}
\end{small}
where $\textbf{H}$ denotes the node embedding matrix, $\tilde{A}=A+I$ where A is the adjacency matrix and $I$ is the identity matrix, $\tilde{D}_{ii}=\sum_j \tilde{A}_{ij}$, and $\textbf{W}$ is the tranable matrix of weights. While the $l$-th layer of GraphSAGE is formulated as:
\begin{small}
\begin{equation}
\begin{aligned}
\textbf{n}_u^{(l)}&=\operatorname{AGG}(\operatorname{ReLU}(\textbf{Q}^{(l)}\textbf{h}_v^{(l)}+\textbf{q}^{(l)}\vert v\in N(u)))\\
\textbf{h}_u^{(l+1)}&=\operatorname{ReLU}(\textbf{W}^{(l)}\operatorname{CONCAT}(\textbf{h}_u^{(l)},\textbf{n}_u^{(l)}))
\end{aligned}
\end{equation}
\end{small}
where $\textbf{h}_u$ denotes the node embedding for node $u$, $N(u)$ denotes the local neighbors of $u$, and $\textbf{Q}$, $\textbf{q}$, $\textbf{W}$ are tainable parameters. The input node features are length-4 one-hot vectors, each dimension of which presents the four node types, i.e., positive literal, negative literals, clauses, and communities.

\subsubsection{Training Loss}
We adopt the binary cross entropy loss to train the classifying GNN modules:
\begin{small}
\begin{align}
    \mathcal{L} = -\mathbb{E}_{u,v\sim p_{pos}}[\log(\sigma(\mathbf{h}_u^\top\mathbf{h}_v)]-\mathbb{E}_{u,v\sim p_{neg}}[\log(1-\sigma(\mathbf{h}_u^\top\mathbf{h}_v)]
\end{align}
\end{small}
where $p_{pos}$ and $p_{neg}$ are the distributions over positive and negative training pairs. GNN modules for in-community and cross-community connection are trained under the same paradigm, while the training data are separately constructed regarding the community affiliation, enabling them to be proficient in different tasks. 

\section{Experiments}
\label{sec:exp}

\subsection{Experimental Setup}

\begin{table}[tb!]
    \centering
    \caption{Statistics of the datasets: $N_v$: number of variables, $N_c$: number of clauses, $T$: CaDiCaL solving time.}
    \vspace{-10pt}
    \resizebox{0.85\linewidth}{!}{
		\begin{tabular}{ccccccc}
    \toprule
	Dataset  &  min. $N_v$ & max. $N_v$   & min. $N_c$ & max. $N_c$ & min. $T$ & max. $T$    \\		\midrule
	MP1 &   400 & 932   & 1737 & 2731 & 55.68 & 410.70\\
	SP4  &  786 & 2052   & 3959 & 5793 & 53.27 & 202.45                \\
         LEC-CORP &  351 & 1577   & 1357 & 6485 & 14.48  & 307.09\\
        \bottomrule
		\end{tabular}
 	}
    \label{tab:data}
    \vspace{-5pt}
\end{table}

The evaluation metrics w.r.t. the augmented formulae include \textit{graph structural statistics} and \textit{computational hardness}. Both metrics are desired to be close to the input benchmarks to achieve the industrial reality of generation. Moreover, we test on a real-world benchmark to show the superiority of our method to help develop better SAT solvers via hyperparameter tuning using our generated new samples, which is a standard protocol both in literature and industry.


\subsubsection{Datasets.} We utilize real-world SAT formulae MP1 and SP4 from SAT Competition 2021~\cite{competition21}, 2022~\cite{competition22}, and industrial circuit instances from a worldwide corporate called LEC-CORP. We divide the data into three datasets corresponding to different scenarios and train the models separately for each dataset, as we desire the model to capture the characteristics of specific data rather than generating formulae fitting the average metrics of the mixed data in general, which may not resemble any scenario. While the formulae in each dataset may have the same logical structure and obey their underlying distribution of hardness. Statistics are given in Table~\ref{tab:data}.\looseness=-1

\textbf{1) MP1 and SP4 datasets from the SAT Competition.} The worldwide competition calls for SAT instances from different challenging problems, two scenarios of which yield these two datasets. The detailed descriptions can be found in \cite{competition21,competition22}.

\textbf{2) LEC-CORP dataset from a corporate.} {Logical Equivalence Checking}~\cite{jansen2003electronic} (LEC) is a classic problem in electronic design automation (EDA), to formally prove that two representations of a circuit design exhibit exactly the same behavior. 
Formal verification of big-bit multipliers has been reported to be one of the important but challenging problems in the verification
community, especially the modern multipliers are often with complex architectures to satisfy the community demands for fast and area-efficient demands. We collect the multipliers, whose bits range from $4 \times 4$ to $22 \times 22$, from industrial LEC scenarios and transform them into CNF formats. This dataset will be used in Sec.~\ref{sec:tuning} to verify the effectiveness of our methods on real-world business data by hyperparameter tuning.\looseness=-1

\subsubsection{Baselines} The evaluation involves traditional hand-crafted formulae generation methods and surging learning-based methods. For hand-crafted methods, Community Attachment (CA)~\cite{giraldez2015modularity} targets at fitting the desired VIG modularity of the generated formulae while Popularity-Similarity (PS)~\cite{giraldez2017locality} tries to fit both the modularity and the power-law distribution of variable occurrences. Learning-based baselines include G2SAT~\cite{you2019g2sat}, and GCN2S, EGNN2S, ECC2S proposed in \cite{garzon2022performance}, which are built upon the node split-merge framework with different network backbones.

\begin{table*}[tb!]	
\centering 
\caption{Results of graph structural statistics of generated formulae. Best match of mean and standard deviation in bold.}
\vspace{-10pt}
\resizebox{0.95\linewidth}{!}{
    \begin{tabular}{ccccccccccc}
    \toprule
    \multirow{2}{*}{Dataset}&\multirow{2}{*}{Method}	& \multicolumn{2}{c}{\textbf{VIG}}&&\multicolumn{2}{c}{\textbf{VCG}}&&\textbf{LIG} &&\textbf{LCG}  \\\cmidrule{3-4}\cmidrule{6-7}\cmidrule{9-9}\cmidrule{11-11}
    
     && Clustering	&  Modularity &&   Clause $\alpha_c$  & Modularity	&&  Modularity  &&  Modularity  \\	\midrule
\multirow{9}{*}{MP1}&
    Ground Truth& 0.32$\pm$0.01 & 0.67$\pm$0.03 &&  3.12$\pm$0.07 & 0.77$\pm$0.03 && 0.79$\pm$0.06 && 0.75$\pm$0.05\\\cmidrule{2-11}
    
    & CA~\cite{giraldez2015modularity}  & 0.15$\pm$0.02 (53.12\%) & 0.57$\pm$0.15 (14.93\%) &&  N/A & 0.67$\pm$0.12 (12.99\%) && 0.58$\pm$0.14 (26.58\%) && 0.50$\pm$0.03 (33.33\%) \\
    
    &PS~\cite{giraldez2017locality} &  0.29$\pm$0.08 (9.38\%) & 0.13$\pm$0.02 (80.60\%) &&  5.46$\pm$0.29 (75.00\%) & 0.28$\pm$0.01 (63.64\%) && 0.17$\pm$0.01 (78.48\%) && 0.29$\pm$0.01 (61.33\%)\\
    
    &G2SAT~\cite{you2019g2sat}  & 0.44$\pm$0.20 (37.50\%) & 0.73$\pm$0.15 (8.96\%) &&  5.09$\pm$0.96 (63.14\%) & 0.82$\pm$0.10 (6.49\%) && 0.87$\pm$0.07 (10.13\%) && 0.68$\pm$0.06 (9.33\%)\\
    
    &GCN2S~\cite{garzon2022performance} & 0.29$\pm$0.11 (9.38\%) & \textbf{0.69$\pm$0.04 (2.99\%)} &&  3.07$\pm$0.22 (1.60\%)& 0.74$\pm$0.05 (3.90\%) && 0.84$\pm$0.01 (6.33\%) && 0.66$\pm$0.03 (12.00\%)\\
    
    &EGNN2S~\cite{garzon2022performance} & 0.41$\pm$0.10 (28.12\%) & 0.70$\pm$0.06 (4.48\%) &&  5.51$\pm$2.25 (76.60\%) & 0.80$\pm$0.09 (3.90\%) && 0.83$\pm$0.05 (5.06\%) && 0.67$\pm$0.05 (10.67\%)\\

    &ECC2S~\cite{garzon2022performance} & 0.20$\pm$0.09 (37.50\%) & 0.53$\pm$0.04 (20.90\%) &&  6.71$\pm$0.18 (115.06\%) & 0.71$\pm$0.01 (7.79\%) && 0.70$\pm$0.03 (11.39\%) && 0.64$\pm$0.01 (14.67\%)\\\cmidrule{2-11}

    &{\mymodel} (GCN)&0.34$\pm$0.03 (6.25\%) & 0.62$\pm$0.03 (7.46\%) &&  \textbf{3.11$\pm$0.19 (0.32\%)} & \textbf{0.74$\pm$0.03 (3.90\%)} && \textbf{0.76$\pm$0.05 (3.80\%)} && 0.69$\pm$0.03 (8.00\%)\\

    &{\mymodel} (SAGE)&\textbf{0.34$\pm$0.01  (6.25\%)} & 0.62$\pm$0.02 (7.46\%)
 &&  4.19$\pm$0.97 (34.29\%) & 0.74$\pm$0.02 (3.90\%) && 0.75$\pm$0.06 (5.06\%) && \textbf{0.69$\pm$0.04 (8.00\%)}\\
    
    \midrule

\multirow{9}{*}{SP4}&
    Ground Truth& 0.57$\pm$0.15 & 0.64$\pm$0.04 &&  4.30$\pm$0.03 & 0.72$\pm$0.03 && 0.78$\pm$0.02 && 0.69$\pm$0.02\\\cmidrule{2-11}
    
    &CA~\cite{giraldez2015modularity}  & 0.20$\pm$0.01 (64.91\%) & 0.58$\pm$0.08 (9.38\%) &&  N/A & 0.74$\pm$0.02 (2.78\%) && 0.61$\pm$0.04 (21.79\%) && 0.58$\pm$0.03 (15.94\%) \\
    
    &PS~\cite{giraldez2017locality} &  0.26$\pm$0.11 (54.39\%) & 0.14$\pm$0.03 (78.12\%) &&  8.38$\pm$0.84 (94.88\%) & 0.28$\pm$0.03 (61.11\%) && 0.13$\pm$0.17 (83.33\%) && 0.28$\pm$0.03 (59.42\%)\\
    
    &G2SAT~\cite{you2019g2sat}  & 0.30$\pm$0.13 (47.37\%) & 0.53$\pm$0.19 (17.19\%) &&  3.49$\pm$0.65 (18.84\%) & 0.70$\pm$0.11 (2.78\%) && 0.65$\pm$0.25 (16.67\%) && 0.66$\pm$0.08 (4.35\%) \\
    
    &GCN2S~\cite{garzon2022performance}& 0.14$\pm$0.01 (75.44\%) & 0.57$\pm$0.13 (10.94\%) && 3.90$\pm$0.09 (9.30\%) & 0.70$\pm$0.08 (2.78\%) && 0.72$\pm$0.15 (7.69\%) && 0.65$\pm$0.05 (5.80\%)\\
    
    &EGNN2S~\cite{garzon2022performance}& 0.22$\pm$0.01 (61.40\%) & 0.54$\pm$0.15 (15.62\%) && 4.79$\pm$1.45 (11.40\%) & 0.72$\pm$0.07  (0.00\%) && 0.69$\pm$0.16 (11.54\%) && 0.65$\pm$0.02 (5.80\%) \\

    &ECC2S~\cite{garzon2022performance}& 0.12$\pm$0.02 (78.95\%) & 0.49$\pm$0.12 (23.44\%) &&  3.91$\pm$0.08 (9.07\%) & 0.70$\pm$0.05 (2.78\%) && 0.66$\pm$0.16 (15.38\%) && 0.64$\pm$0.02 (7.25\%)\\\cmidrule{2-11}

    &{\mymodel} (GCN)& \textbf{0.56$\pm$0.15 (1.75\%)} & 0.65$\pm$0.03 (1.56\%) &&  \textbf{4.12$\pm$0.02 (4.19\%)} & \textbf{0.72$\pm$0.03 (0.00\%)} && \textbf{0.78$\pm$0.02 (0.00\%)} && 0.67$\pm$0.03 (2.90\%)\\

    &{\mymodel} (SAGE)& 0.56$\pm$0.13 (1.75\%) & \textbf{0.65$\pm$0.04 (1.56\%)} &&  3.75$\pm$0.44 (12.79\%) & 0.71$\pm$0.02 (1.39\%) && \textbf{0.78$\pm$0.02 (0.00\%)} && \textbf{0.67$\pm$0.02 (2.90\%)}\\
    \midrule

\multirow{9}{*}{LEC-CORP}&
    Ground Truth& 0.40$\pm$0.01 & 0.65$\pm$0.02 &&  6.29$\pm$0.02 & 0.78$\pm$0.02 && 0.70$\pm$0.02 && 0.66$\pm$0.02\\\cmidrule{2-11}
    
    &CA~\cite{giraldez2015modularity} &  0.21$\pm$0.03 (47.50\%) & 0.62$\pm$0.03 (4.62\%) &&  N/A & 0.74$\pm$0.01 (5.13\%) && 0.64$\pm$0.02 (8.57\%) && 0.54$\pm$0.01 (18.18\%)\\
    
    &PS~\cite{giraldez2017locality}&  0.26$\pm$0.07 (35.00\%) & 0.13$\pm$0.01 (80.00\%) &&  8.03$\pm$0.71 (27.66\%) & 0.28$\pm$0.09 (64.10\%) && 0.15$\pm$0.00 (78.57\%) && 0.28$\pm$0.00 (57.58\%)\\
    
    &G2SAT~\cite{you2019g2sat}  & 0.34$\pm$0.10 (15.00\%) & 0.81$\pm$0.01 (24.62\%) &&  6.80$\pm$0.91 (8.11\%) & 0.85$\pm$0.02 (8.97\%) && 0.91$\pm$0.01 (30.00\%) && 0.68$\pm$0.02 (3.03\%)\\
    
    &GCN2S~\cite{garzon2022performance}& 0.26$\pm$0.06 (35.00\%) & 0.76$\pm$0.02 (16.92\%) &&  2.98$\pm$0.01 (52.62\%) & 0.81$\pm$0.01 (3.85\%) && 0.88$\pm$0.01 (25.71\%) && \textbf{0.65$\pm$0.01 (1.52\%)}\\
    
    &EGNN2S~\cite{garzon2022performance} & 0.25$\pm$0.09 (37.50\%) & 0.36$\pm$0.01 (44.62\%) &&  4.86$\pm$1.78 (22.73\%) & 0.61$\pm$0.02 (21.79\%) && 0.59$\pm$0.01 (15.71\%) && 0.58$\pm$0.02 (12.12\%)\\

    &ECC2S~\cite{garzon2022performance}& 0.13$\pm$0.02 (67.50\%) & 0.27$\pm$0.01 (58.46\%) &&  6.78$\pm$1.16 (7.79\%) & 0.54$\pm$0.01 (30.77\%) && 0.49$\pm$0.01 (30.00\%) && 0.55$\pm$0.01 (16.67\%)\\\cmidrule{2-11}

    &{\mymodel} (GCN) & 0.42$\pm$0.02 (5.00\%) & 0.69$\pm$0.03 (6.15\%) &&  4.39$\pm$1.30 (30.21\%) & \textbf{0.78$\pm$0.02 (0.00\%)} && 0.71$\pm$0.03 (1.43\%) && 0.64$\pm$0.01 (3.03\%)\\
    &{\mymodel} (SAGE)& \textbf{0.40$\pm$0.01 (0.00\%)} & \textbf{0.66$\pm$0.02 (1.54\%)} &&  \textbf{6.42$\pm$1.16 (2.07\%)} & 0.78$\pm$0.01 (0.00\%) && \textbf{0.69$\pm$0.02 (1.43\%)} && 0.64$\pm$0.01 (3.03\%) \\
    
    \bottomrule
    \end{tabular}
}
\label{tab:structure}
\vspace{-5pt}
\end{table*}

\begin{figure}
    \centering
    \includegraphics[width=0.98\linewidth]{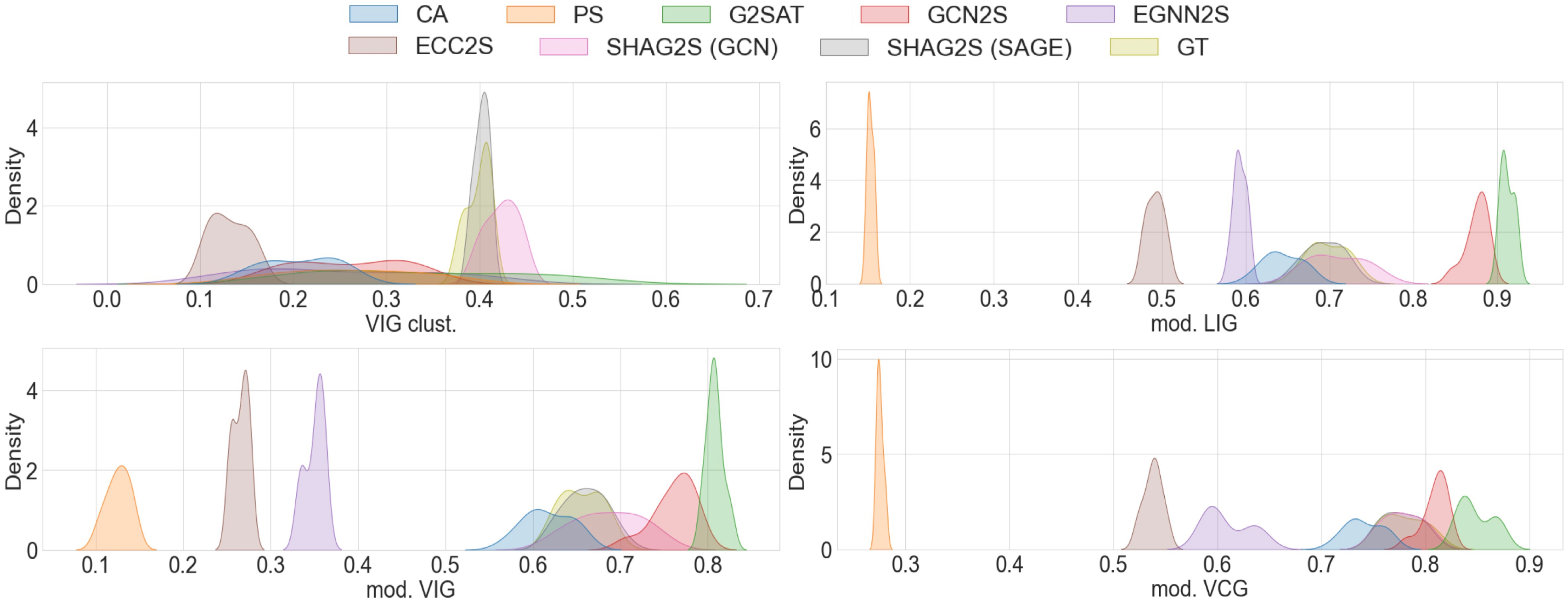}
    \vspace{-10pt}
    \caption{Visualization of graph statistics distributions of generated formulae from the private LEC-CORP dataset.}
    \label{fig:stats}
    \vspace{-5pt}
\end{figure}

\subsection{Performance on Structural Properties}

Benchmarks from different practical applications are actually encoding different problems and obey different distributions in structures. The generated formulae are desired to exhibit similar structural properties with the original input formulae in training. 

\subsubsection{Metrics} We follow previous studies~\cite{ansotegui2009structure,newsham2014impact,you2019g2sat,garzon2022performance} to evaluate the structural properties by graph statistics. The metrics include modularity~\cite{newman2004finding} in VIG, VCG, LIG, LCG, averaging clustering coefficient~\cite{newman2001clustering} in VIG, and the scale-free structure parameters of clause $\alpha_c$~\cite{ansotegui2009structure,clauset2009power} in VCG. We omit the scale-free metric of the variable as it remains unaltered in the graph templates.

\subsubsection{Results} The quantitative results are presented in Table~\ref{tab:structure}. As seen, hand-crafted methods, including CA and PS, can fit well some of the statistics while failing to perform well on other metrics. Learning-based methods generally capture the global structural properties and are able to plausibly resemble all the graph statistics compared to hand-crafted generation methods. However, there are still margins left for previously existing methods to improve the statistic approximation, while {\mymodel} on top of GCN and SAGE backbones outperforms all the peer methods by a notable margin. The calculation of the graph statistics is enforced on instances generated for each datasets and we present the mean and the standard deviation of the statistic distribution. To more directly illustrate the metric difference to the original benchmarks, we compute the relative error regarding the ground truth listed with the metrics. In general, compared to the best results achieved by previous methods, we reduce the relative error by \textbf{38.5\% on average}. The performance gains are more significant on more complex datasets, e.g., SP4 and LEC. The most notable result is achieved in VIG clustering measure on SP4 dataset, where previous methods cannot successfully resemble the statistic and merely achieve the best 47.37\% relative error, while {\mymodel} reduces it to 1.75\%.

We plot the graph statistic distributions of the generated formulae in Fig.~\ref{fig:stats}. We select  LEC-CORP data for visualization since it maintains the most instances to better estimate the probability. We show the estimated probability density of the graph statistics at different values, which is smoothed by a kernel density estimator. As seen, {\mymodel} fits the best to the ground truth benchmarks on all graph statistics, while baseline methods fail to produce a plausible fit on the overall measures.

\subsection{Performance on Hardness Reproducing}
This section evaluates the computational hardness of the generated formulae w.r.t. the original benchmarks. As the desire raised by the ten key challenges in propositional reasoning and search~\cite{kautz1997ten}, the generated formulae shall maintain similar computational properties with the original benchmarks as closely as possible.

\subsubsection{Metrics.} We directly measure the hardness of the generated formulae through SAT solver runtime as well as the ranking of solver performance following~\cite{you2019g2sat}. For solver ranking evaluation, we select mainstream solvers Kissat, CaDiCaL and Glucose and rank them based on the performance over the generated formulae. We evaluate how the ranking over generated data aligns with that of the training data and we also report the ratio of instances that exactly recover the ranking. For runtime evaluation, we use CaDiCaL~\cite{fleury2020cadical} as the test solver and utilize the runlim tool (http://fmv.jku.at/runlim/) to measure the solving time for each instance. Since the solving time of different ground truth formulae can vary much and can cause large standard deviations, we report results for each ground truth instance, i.e., generating and evaluating the formulae based on the template produced by each ground truth instance.

\begin{table}[tb!]
    \centering
    \caption{Relative SAT solver performance on GT formulae and generated  formulae. $S_1$: Kissat; $S_2$: CaDiCaL; $S_3$: Glucose.}
    \vspace{-10pt}
    \resizebox{0.7\linewidth}{!}{
\begin{tabular}{ccccc}
    \toprule

Method & Average Performance Ranking  & Accuracy \\\midrule

    Groud Truth & ($S_1$, $S_2$, $S_3$) & N/A  \\\midrule
    
    CA~\cite{giraldez2015modularity} & \underline{($S_1$, $S_2$, $S_3$)} & 37.5\% \\
    
    PS~\cite{giraldez2017locality}& ($S_2$, $S_1$, $S_3$) & 25.0\% \\
    
    G2SAT~\cite{you2019g2sat}  & ($S_1$, $S_3$, $S_2$) & 25.0\% \\
    
    GCN2S~\cite{garzon2022performance}& ($S_2$, $S_1$, $S_3$) & 37.5\% \\
    
    EGNN2S~\cite{garzon2022performance}& ($S_3$, $S_1$, $S_2$) & 37.5\% \\

    ECC2S~\cite{garzon2022performance}&  ($S_2$, $S_1$, $S_3$) & 25.0\% \\


    {\mymodel}& \underline{($S_1$, $S_2$, $S_3$)} & \textbf{100.0\%} \\

    \bottomrule
		\end{tabular}
 	}
    \label{tab:rank}
    \vspace{-5pt}
\end{table}

\begin{table*}[tb!]
    \centering
    \caption{CPU solving runtime evaluation. The backbones for our method are in brackets. Best match with GT are in bold.}
    \vspace{-10pt}
    \resizebox{0.98\linewidth}{!}{
\begin{tabular}{cc|cccccc|cc|c}
    \toprule

Dataset & Instance  & CA~\cite{giraldez2015modularity}& PS~\cite{giraldez2017locality}& G2SAT~\cite{you2019g2sat} & GCN2S~\cite{garzon2022performance}& EGNN2S~\cite{garzon2022performance} & ECC2S~\cite{garzon2022performance} & {\mymodel} (GCN) & {\mymodel} (SAGE) & Ground Truth\\\midrule

\multirow{3}{*}{MP1}
&mp1-1 & 0.22 $\pm$ 0.14  (99.61\%) & 0.77 $\pm$ 0.36 (98.62\%) & 0.25 $\pm$ 0.09 (99.55\%) & 0.80 $\pm$ 0.12 (98.57\%) & 0.64 $\pm$ 0.08 (98.85\%)& 0.71 $\pm$ 0.07 (98.73\%) & 46.01 $\pm$ 9.81 (17.37\%) & \textbf{53.29 $\pm$ 12.71 (4.30\%)} & 55.68\\
&mp1-2  & 568.90 $\pm$ 705.65 (893.88\%) & 0.37 $\pm$ 0.12 (99.35\%) & 0.70 $\pm$ 0.11 (98.77\%) & 0.48 $\pm$ 0.09 (99.16\%) & 0.61 $\pm$ 0.45 (98.93\%)& 0.25 $\pm$ 0.08 (99.56\%) & 41.58 $\pm$ 13.48 (27.36\%) & \textbf{50.34 $\pm$ 14.47(12.05\%)} & 57.24\\
&mp1-3  & 0.30 $\pm$ 0.07 (99.93\%) & 0.56 $\pm$ 0.12 (99.86\%)& 0.69 $\pm$ 0.41 (99.83\%) & 0.35 $\pm$ 0.38 (99.91\%) & 0.42 $\pm$ 0.23 (99.89\%)& 0.40 $\pm$ 0.18 (99.90\%) & \textbf{538.772 $\pm$ 196.21 (31.18\%)} & 578.51 $\pm$ 130.60 (40.86\%) & 410.7\\\midrule

\multirow{3}{*}{SP4}
&sp4-1 & 0.44$\pm$0.29 (99.17\%) & 0.57$\pm$0.29 (98.93\%) & 0.65$\pm$0.10 (98.78\%) & 0.30$\pm$0.12 (99.44\%) & 0.86$\pm$0.13 (98.39\%)& 0.65$\pm$0.10 (98.78\%) & 39.76$\pm$8.41 (25.36\%) & \textbf{54.10$\pm$4.81 (1.55\%)} & 53.27\\
&sp4-2  & 0.61$\pm$0.28 (99.70\%) & 0.46$\pm$0.27 (99.77\%) & 0.61$\pm$0.38 (99.70\%) & 0.54$\pm$0.12 (99.73\%) & 0.66$\pm$0.38 (99.67\%)& 0.65$\pm$0.10 (99.68\%) & \textbf{119.54$\pm$ 12.09 (40.90\%)} & 113.47$\pm$ 14.30 (43.90\%) & 202.45\\
&sp4-3  &0.57$\pm$0.27 (99.28\%)& 0.48$\pm$0.27 (99.39\%)& 0.32$\pm$0.10 (99.60\%) & 0.92$\pm$0.11 (98.84\%) & 0.41$\pm$0.12 (99.48\%)& 0.65$\pm$0.10 (99.18\%) & \textbf{64.12$\pm$6.37 (19.02\%)} & 56.12$\pm$3.89 (29.12\%) & 79.18\\\midrule

\multirow{8}{*}{LEC-CORP}
&lec-1  & 0.60$\pm$0.31 (98.64\%) & 0.63$\pm$0.26 (98.57\%) & 0.33$\pm$0.27 (99.25\%) & 0.28$\pm$0.22 (99.36\%) & 0.57$\pm$0.26 (98.70\%) & 0.80$\pm$0.07 (98.18\%) & 46.05$\pm$3.75 (4.75\%) & \textbf{45.40$\pm$3.46 (3.27\%)} & 43.96\\
&lec-2 & 0.52$\pm$0.32 (96.41\%) & 0.44$\pm$0.30 (96.96\%) & 0.60$\pm$0.30 (95.86\%) & 0.50$\pm$0.25 (96.55\%) & 0.57$\pm$0.30 (96.06\%) & 0.72$\pm$0.27 (95.03\%) & \textbf{15.45$\pm$0.33 (6.70\%)} & 15.88$\pm$0.99 (9.67\%) & 14.48\\
&lec-3  & 1.66$\pm$0.53 (98.55\%) & 0.50$\pm$0.27 (99.56\%)  & 0.84$\pm$0.06 (99.26\%) & 0.36$\pm$0.22 (99.68\%) & 0.52$\pm$0.26 (99.54\%) & 0.70$\pm$0.31 (99.39\%) & \textbf{106.04$\pm$3.32 (7.10\%)} & 95.24$\pm$4.78 (16.57\%) & 114.15\\
&lec-4  & 20.73$\pm$30.04 (64.38\%) & 0.39$\pm$0.29 (99.33\%) & 0.34$\pm$0.22 (99.42\%) & 0.42$\pm$0.22 (99.28\%) & 0.47$\pm$0.34 (99.19\%) & 0.59$\pm$0.18 (98.98\%) & 60.40$\pm$2.87 (3.78\%) & \textbf{56.57$\pm$2.89 (2.80\%)} & 58.20\\
&lec-5  & 0.75$\pm$0.18 (98.76\%) & 0.49$\pm$0.16 (99.19\%) & 0.63$\pm$0.31 (98.96\%) & 0.63$\pm$0.22 (98.96\%) & 0.30$\pm$0.25 (99.50\%) & 0.35$\pm$0.31 (99.42\%) & 53.10$\pm$3.81 (12.04\%) & \textbf{58.34$\pm$10.97 (3.36\%)} & 60.37\\
&lec-6  & 0.64$\pm$0.30 (99.79\%) & 0.60$\pm$0.24 (99.80\%) & 0.56$\pm$0.30 (99.82\%) & 0.47$\pm$0.34 (99.85\%) & 0.51$\pm$0.23 (99.83\%) & 0.64$\pm$0.27 (99.79\%) & \textbf{310.51$\pm$35.82 (1.11\%)} & 319.19$\pm$55.38 (3.94\%) & 307.09\\
&lec-7  & 152.42$\pm$224.82 (355.26\%) & 0.37$\pm$0.23 (98.89\%) & 0.67$\pm$0.12 (98.00\%) & 0.54$\pm$0.32 (98.39\%) & 0.64$\pm$0.31 (98.09\%) & 0.51$\pm$0.26 (98.48\%) & 37.75$\pm$1.61 (12.75\%) & \textbf{35.32$\pm$1.56 (5.50\%)} & 33.48\\
&lec-8  & 140.81$\pm$142.26 (39.17\%) & 0.36$\pm$0.28 (99.64\%) & 0.61$\pm$0.25 (99.40\%) & 0.57$\pm$0.25 (99.44\%) & 0.53$\pm$0.24 (99.48\%) & 0.37$\pm$0.10 (99.63\%) & 120.98$\pm$7.90 (19.57\%) & \textbf{109.07$\pm$5.18 (7.80\%)} & 101.18\\

    \bottomrule
\end{tabular}
 	}
    \label{tab:hardness}
    \vspace{-5pt}
\end{table*}

\begin{table}[tb!]
    \centering
    \caption{Results of SAT solver hyperparameter tuning to enhance its performance to specific benchmarks. $T_g$ denotes the average runtime of generated instances. $T_t$ denotes the average solving time of the test instances.}
    \vspace{-10pt}
    \resizebox{0.92\linewidth}{!}{
\begin{tabular}{cccccc}
    \toprule

Dataset &Method & Best Hyperparameters  & $T_g$ (s) & $T_t$ (s) & Gain \\\midrule

\multirow{8}{*}{MP1}&
    Default & (0.8, 0.999) & N/A & 642.056 & N/A \\\cmidrule{2-6}
    
    & CA~\cite{giraldez2015modularity} & (0.999, 0.99) & 151.347 & 698.469 & -8.79\%\\
    
    &PS~\cite{giraldez2017locality}& (0.9, 0.999) & 0.021 & 582.874 & +9.22\%\\
    
    &G2SAT~\cite{you2019g2sat}  & (0.85, 0.85) & 0.004 & 583.954 & +9.05\%\\
    
    &GCN2S~\cite{garzon2022performance}& (0.8, 0.9) & 0.011 & 613.705 & +4.42\%\\
    
    &EGNN2S~\cite{garzon2022performance}& (0.75, 0.75) & 0.011 & 624.483 & +2.74\%\\

    &ECC2S~\cite{garzon2022performance}&  (0.75, 0.75) & 0.011 & 624.483 & +2.74\% \\


    &{\mymodel}& (0.95 , 0.8) & 264.994 & \textbf{499.583} & \textbf{+22.19\%}\\\midrule

    \multirow{8}{*}{LEC-CORP}
    & Default & (300, 2, 950) & N/A & 62.344 & N/A\\\cmidrule{2-6}
    
    & CA~\cite{giraldez2015modularity} & (10, 1998046439, 502) & 3.960 & 66.718 & -7.02\%\\
    
    &PS~\cite{giraldez2017locality}& (210245, 772680448, 988) & 0.026 & 151.810 & -143.50\%\\
    
    &G2SAT~\cite{you2019g2sat}  & (455598, 1998371347, 993) & 0.028 & 148.557& -138.29\%\\
    
    &GCN2S~\cite{garzon2022performance}&  (955172, 586177088, 728) & 0.027 & 160.313 & -157.14\%\\
    
    &EGNN2S~\cite{garzon2022performance}& (570715, 598666048, 770) & 0.026 & 158.462& -154.17\%\\

    &ECC2S~\cite{garzon2022performance}&  (435325, 86581488, 604) & 0.027 & 157.609 & -152.81\%\\

    &{\mymodel} & (10, 1999968040, 999) & 50.074 & \textbf{52.972}& \textbf{+15.03\%}\\


    \bottomrule
		\end{tabular}
 	}
    \label{tab:tuning}
    \vspace{-5pt}
\end{table}

\subsubsection{Results.} Table~\ref{tab:rank} shows the solver ranking results on the private LEC-CORP dataset. Correct recoveries of solver ranking to the ground truth are indicated by underlining. {\mymodel} is the only method achieving the exact ground truth ranking with 100\% accuracy. CaDiCaL's runtime is presented in Table~\ref{tab:hardness}. The specific results for each instance are given in Appendix~\ref{app:more_rank}. Note that the previous methods, including both hand-crafted and learning-based methods, all fail to reproduce plausible computational hardness compared to the original benchmarks, which generally generate instances that can be solved in quite a short time. While the generated instances of {\mymodel} can maintain similar  performance to the original ones. The small variance to the ground truth implies that the generated instances  potentially lie in the same distribution of the computational hardness. We again compute the relative error regarding the GT listed with the runtime. Compared to the best results by the previous methods, our method reduces the relative error by \textbf{88.4\%} on average, which appears to be the only method in successfully generating formulae that maintain similar computational hardness to the original ones. Runtime evaluation of more solvers, e.g. Kissat and Glucose are given in Appendix~\ref{app:more_runtime}.

\subsection{Structural Control Ability Study}
\begin{figure}[tb!]
    \centering 
    \subfigure[VCG Modularity]{\includegraphics[width=0.32\linewidth]{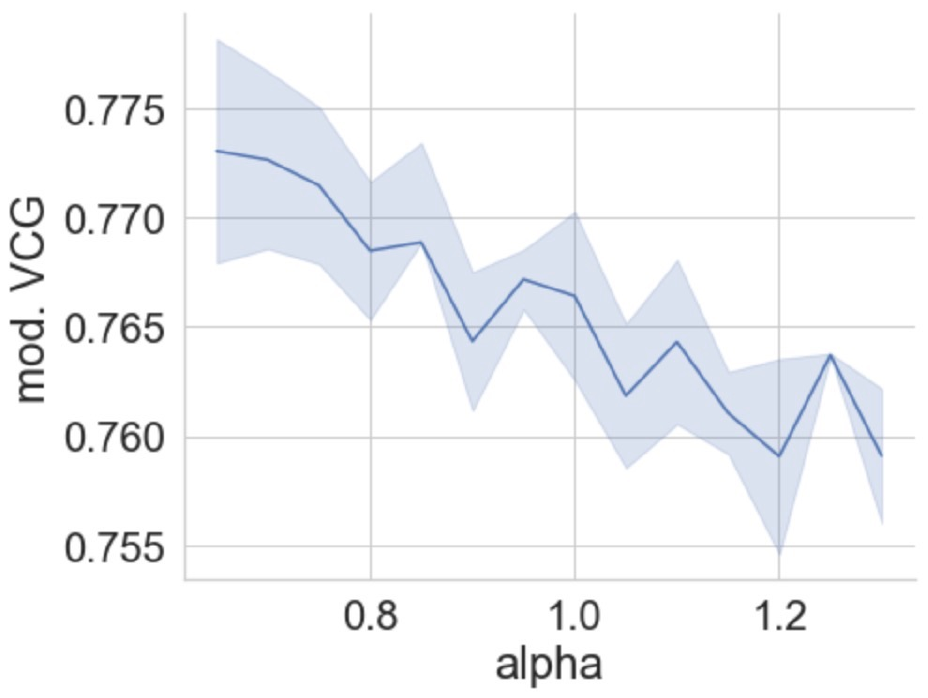}}
    \subfigure[LIG Modularity] {\includegraphics[width=0.32\linewidth]{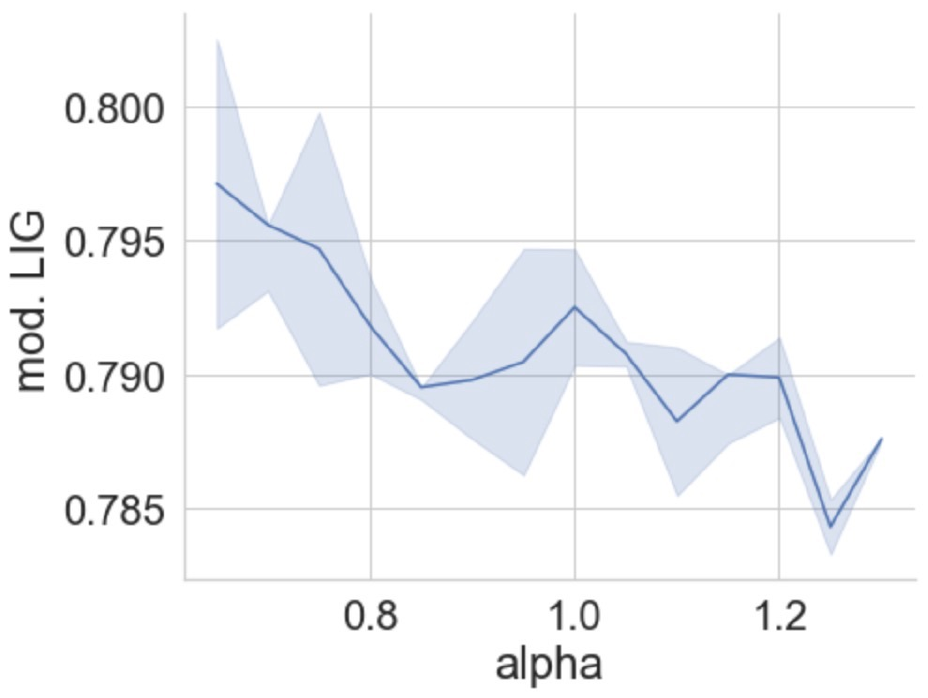}}
    \subfigure[LCG Modularity] {\includegraphics[width=0.32\linewidth]{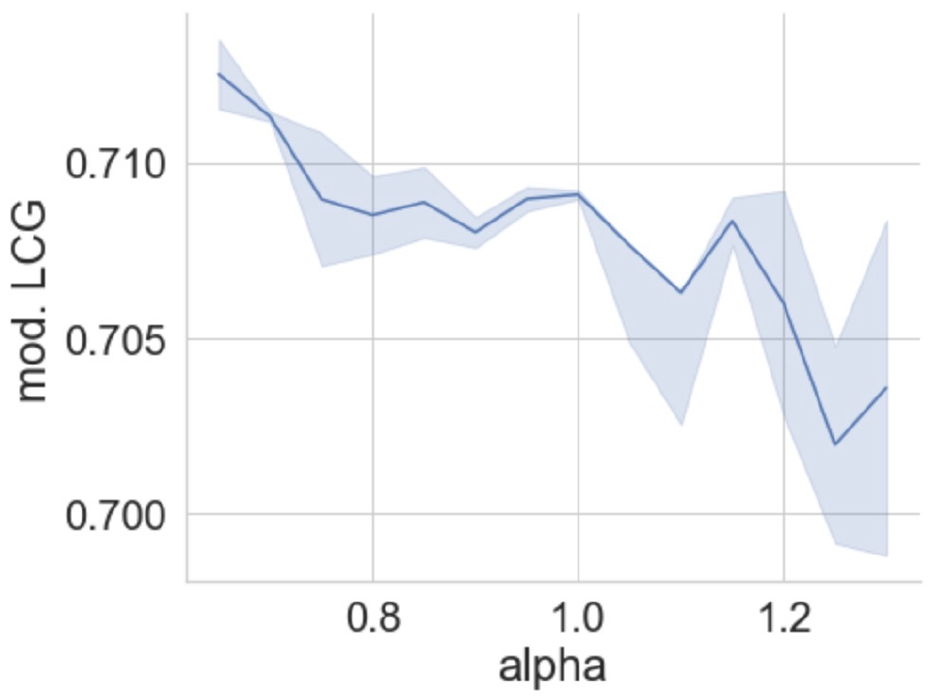}}
    \vspace{-10pt}
    \caption{Effect of $\alpha$ to the modularity of VCG, LIG and LCG on MP1 dataset. Shadow for 30\% confidence interval.}
    \label{fig:alpha}
    \vspace{-5pt}
\end{figure}

As described in Sec.~\ref{sec:cmty-connect}, the community-based phase division and the tunable phase transition point enable {\mymodel} to control the graph structural statistics of the generated formulae within a certain margin. Here we verify the effectiveness of {\mymodel} to control the structural properties by the phase transition hyperparameter $\alpha$. Fig.~\ref{fig:alpha} shows the modularity variation on VCG, LIG and LCG representations while varying $\alpha$ from 0.65 to 1.30 by interval 0.05. As seen, a larger $\alpha$  produces graphs with smaller modularity.

\subsection{Shipping SAT Solvers via Hyperparameter Tuning based on Generated Instances}
\label{sec:tuning}
SAT solvers with default hyperparameter settings generally perform well on major benchmarks. But in practice, solvers tailored for specific applications are needed. Here we further show an application of generated instances to tune hyperparameters of SAT solvers and thus develop solvers tailored for specific benchmarks. Here we tune hyperparameters of mainstream SAT solvers to develop tailored solvers for specific real-world problems. The verified solvers include Glucose~\cite{audemard2009glucose} tuned on MP1 dataset and more powerful CaDiCaL~\cite{fleury2020cadical} tuned on  LEC-CORP dataset, which is deployed in a worldwide major company's EDA scenario to solve LEC problems.

\subsubsection{Experimental Settings.} For each method, we augment the data five times as the original benchmarks to guide the hyperparameter selection of SAT solvers. For MP1 dataset, we conduct a grid search over two of Glucose's hyperparameters, i.e., variable decay and clause decay, which determine the ordering of variables to be searched and the ordering of the learned clauses to be removed respectively. The grid search is conducted over \{0.75, 0.8, 0.85, 0.9, 0.95, 0.99, 0.999\} for both variable decay and clause decay, including in total 49 combinations of hyperparameter selections. We measure the solver runtime and select the best hyperparameters for the test. For \textbf{LEC-CORP} dataset from practical industrial scenarios, we adopt one of the state-of-the-art Bayesian optimization tool HEBO~\cite{Cowen-Rivers2022-HEBO} to tune more powerful solver CaDiCaL, namely reduce interval, restart interval, and score factor, which pose impacts on the frequency of learned clause reduction, frequency of restarting, and the variable scoring policy respectively. The tuning range of these three hyperparameters are [10, 1e6], [1, 2e9], and [500, 1e3] respectively. The solver runtime limit is set to 1,000 seconds.

\subsubsection{Results.}
We report the best hyperparameter tuple, the average runtime of the generated instances, and the average runtime of the test instances in Table ~\ref{tab:tuning}. The performance gain is calculated to reflect the runtime reduction rate after tuning on different sets of generated data. For MP1, {\mymodel} achieves the best performance gain of 22.19\%, significantly surpassing the best performance of previous methods by \textbf{140.7\%}. While for complex  LEC-CORP data, we are the only method achieving a positive performance gain, lifting the best performance \textbf{from -7.02\% to +15.03\%}.

\section{Conclusion}
This work firstly shows the limitation of previous learning-based methods in reproducing the hardness can presumably stem from the inherent homogeneity in the single-stage split-merge procedure. Then we propose a structure and hardness aware SAT formula generation pipeline with multi-stage, which poses a fine-grained control mechanism to better mimic SAT formulae's structural and computational properties. Empirical results show its superiority over state-of-the-art learning-based and hand-crafted methods.


\begin{acks}
The work was supported in part by China Key Research and Development Program (2020AAA0107600), NSFC (62222607), STCSM (22511105100), SJTU Scientific and Technological Innovation Funds.
\end{acks}

\clearpage
\normalem
\bibliographystyle{ACM-Reference-Format}
\balance
\bibliography{ref}
\newpage
\clearpage
\clearpage
\newpage
\appendix
\section*{Appendix}

\section{Implementation Details}

\subsection{Details of the Model including Hyper-parameter Setting}

For the graph networks, we adopt the 3-layer GCN and SAGE to learn the structures. The features of the nodes are extracted by a linear layer from a length-4 one-hot vector indicating the node types to a length-32 vector. The hidden dimension and the output dimension of the graph networks are both 32. 

For training settings, we train each model for 200 epochs with a learning rate of 0.001. For inference settings, each node pair to be merged  is selected from 100 sample candidates, where the selected pair maintains the largest predicted probability to be merged. $\alpha$ is set as 1.2 for MP1 dataset, and 0.9 for SP4 and LEC-CORP datasets.\looseness=-1

\subsection{Details of Community-based Phase Division}

In the implementation, the community information is embedded into the LCG graph representation in two folds: 1) community nodes as described in Sec.~\ref{sec:build}; 2) an additional community affiliation list indicating the community id of variable nodes. The community affiliation list enables the model to accelerate operations w.r.t. the community structure. For example, while producing train data for cross-community connections, {\mymodel} traverses the clause nodes to check the involved communities of variables in the clause. Through the affiliation list, with the 1-hop neighbor list, it only needs one query operation to obtain the involved communities.


While splitting the formulae, two sets of training data are formed sequentially for cross-community and in-community connection. The two graph networks capturing different probability distributions of cross-community and in-community connection can then be trained in parallel on top of the two data sets. While generating the formulae, the two networks sequentially guide the node merging operations as an inverse process of the splitting phase.

\subsection{Details of Core Scrambling Algorithm}
\label{app:scramble}

Specifically, performing the core scrambling requires core files in DIMACS format corresponding to the formulae which in the implementation are detected by SAT solver CaDiCal~\cite{fleury2020cadical} and DRAT-trim checker~\cite{wetzler2014drat}. The core file is specifically obtained by first solving formulae with CaDiCal to get drat proof file and then analyzing solving process revealed by drat proof file with the DRAT-trim checker. The checker will output the unsatisfiable core which is also in DIMACS format. Note that the unsatisfiable core is not unique, which can be different depending on the solvers' algorithm and implementation. Nevertheless, the cores detected in the industrial instances with recognizable solving hardness always show a major impact in terms of computational hardness. The scrambled core will be output in DIMACS format, denoted as $K_{origin}$.



\subsection{Details of Post-processing}

Post-processing is conducted by iteratively loosing unexpected small cores. At step $i$, the algorithm is fed the last step's output formula $F_{i-1}$ and output $F_i$. Denoting the unsatisfiable core as $K_{i-1}$, together with the original core $K_{origin}$, the algorithm identifies clauses in $ K_{i-1}\setminus K_{origin}$, which are excluded by $K_{origin}$ and included by the unexpected small core $K_{i-1}$. To maintain the dominant role of $K_{origin}$ in computational properties, as mentioned in Sec.~\ref{subsec:core}, we arbitrarily pick a clause in $K_{i-1}\setminus K_{origin}$ and add a new variable to loose $K_{i-1}$. The iteration will stop once the current formula hits the required hardness threshold measured by the solving time of SAT solvers. The threshold can be tuned manually to fit with the original formula as well as the reference benchmark.

\subsection{CNF Representation}

The raw CNF formulae in Sec.~\ref{sec:related} are represented in DIMACS standard format in this work, which is defined by DIMACS (Center for Discrete Mathematics and Theoretical Computer Science) for representing undirected graphs. It has been widely applied in several DIMACS Computational Challenges and is also the CNF format adopted in yearly SAT competitions~\cite{competition22}. The DIMACS format is of textual form and usually appears with the suffix '.cnf'. The file begins with header lines started with 'c' or 'p', which respectively tell comments or numbers of variables and clauses. Each line following the headers is a clause, i.e., a disjunction of literals. Variables are represented by integers (starting from 1) and their negation. 

\section{Experiment Details}

\subsection{Evaluation}

The modularities are calculated by the Louvain Algorithm~\cite{blondel2008fast} implemented by \cite{aynaud2018community}. The estimate of $\alpha_v$ and $\alpha_c$ is calculated by the maximum likelihood method implemented by \cite{ansotegui2009structure}.

\subsection{Computing Facilities}
Experiments on MP1 and SP4 datasets are generally performed on a single GPU of GeForce RTX 3090. While experiments on LEC-CORP from the industrial applications are performed on Tesla V100.

\section{More Experimental Results}




\subsection{Learning Curves}

\begin{figure*}[tb!]
    \centering 
    \subfigure[Loss]{\includegraphics[width=0.19\textwidth]{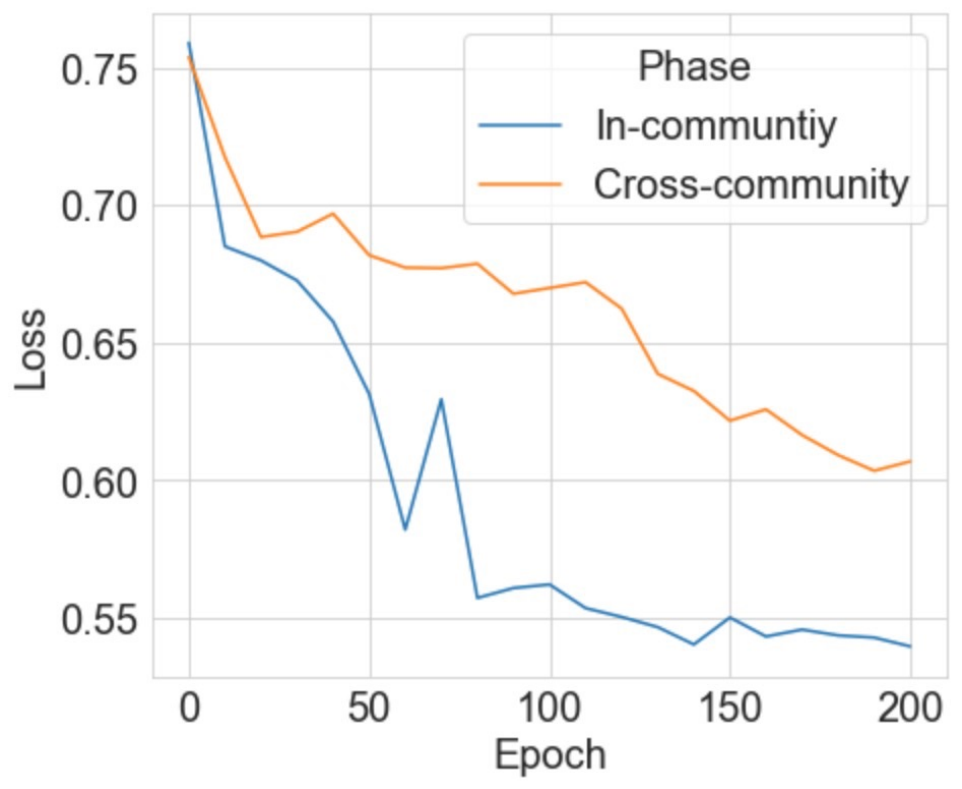}}
    \subfigure[Train AUC] {\includegraphics[width=0.19\textwidth]{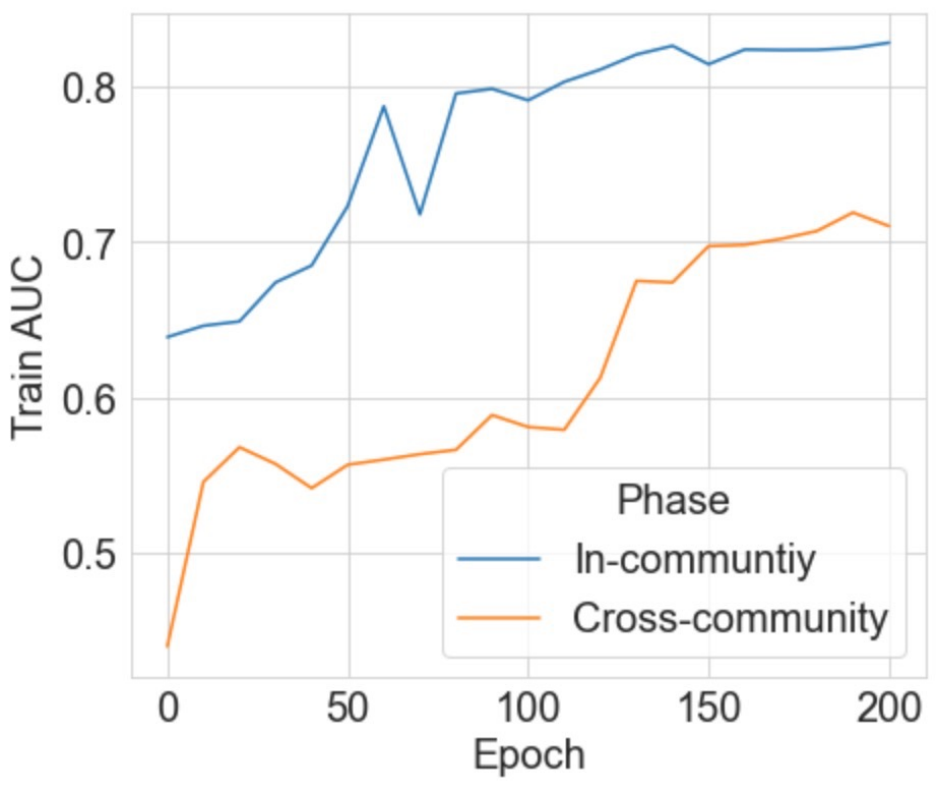}}
    \subfigure[Train ACC] {\includegraphics[width=0.19\textwidth]{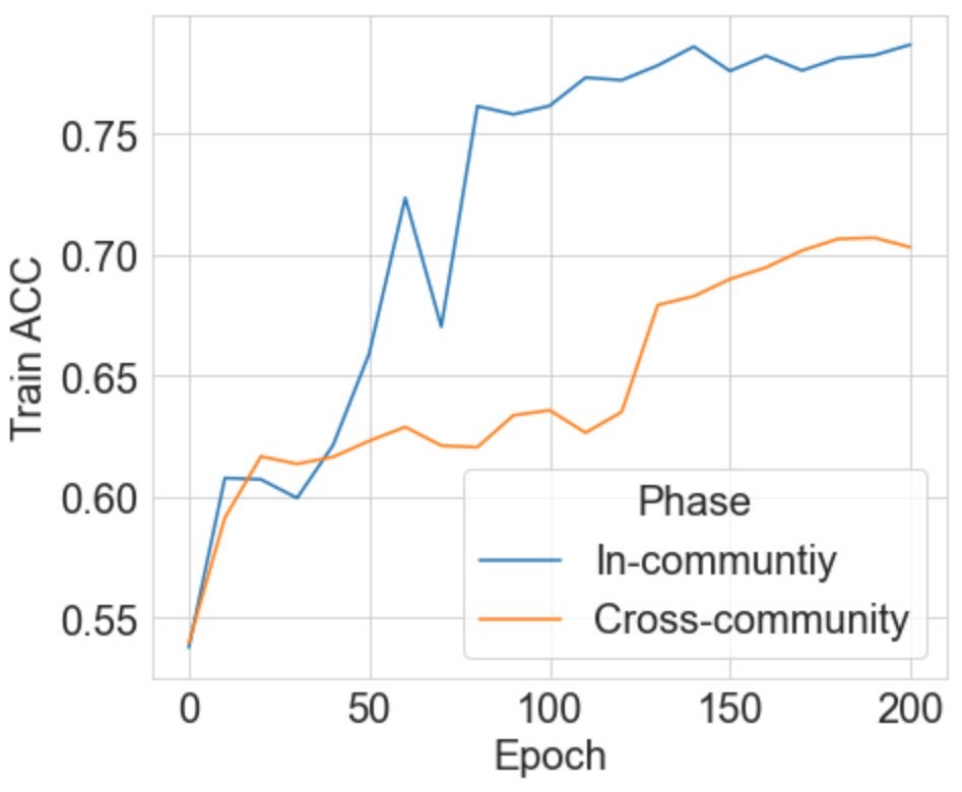}}
    \subfigure[Test AUC] {\includegraphics[width=0.19\textwidth]{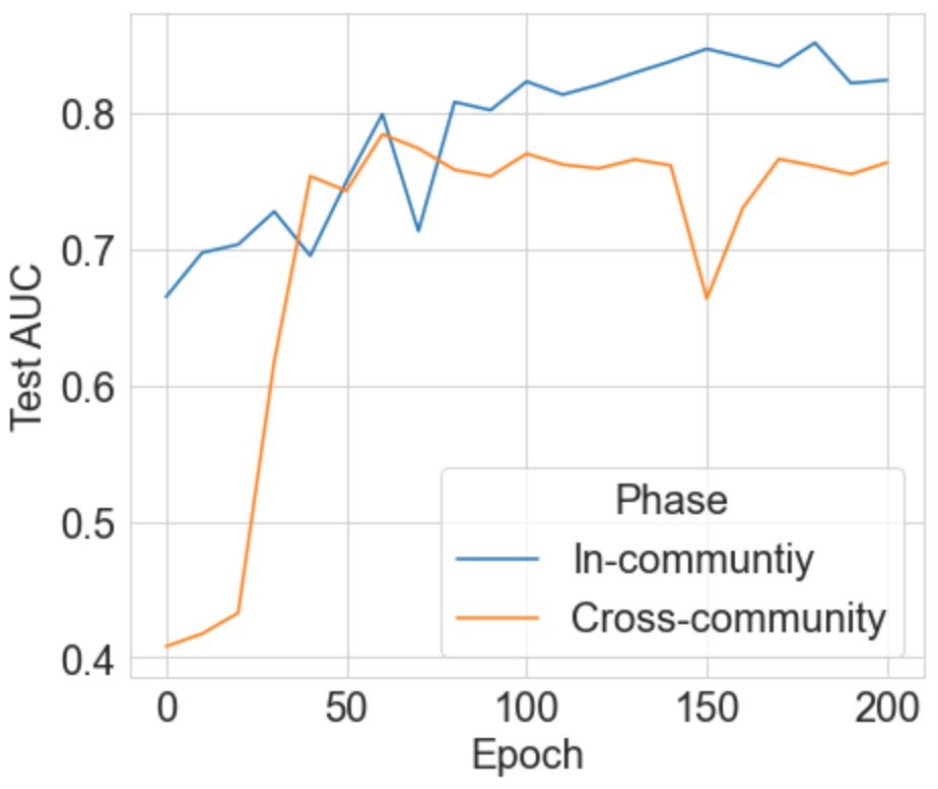}}
    \subfigure[Test ACC] {\includegraphics[width=0.19\textwidth]{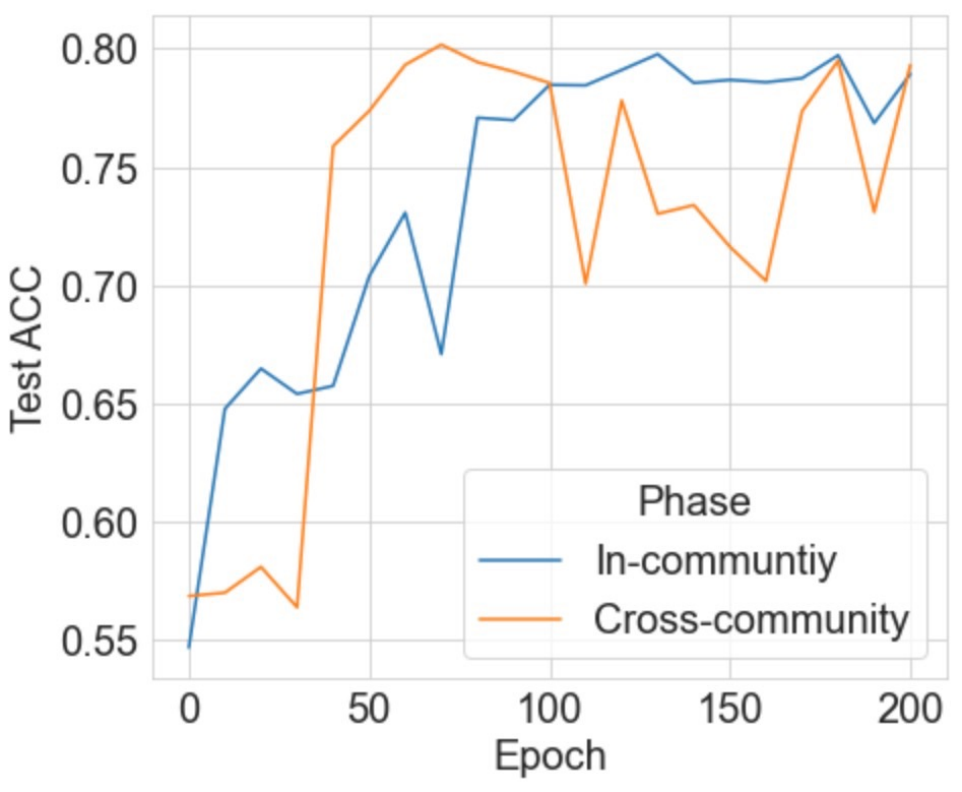}}
    \vspace{-10pt}
    \caption{Learning curves of {\mymodel} with SAGE backbone on MP1 datasets.}
    \label{fig:curves}
    \vspace{-5pt}
\end{figure*}

\begin{figure*}[tb!]
    \centering 
    \subfigure[CA]{\includegraphics[width=0.19\linewidth]{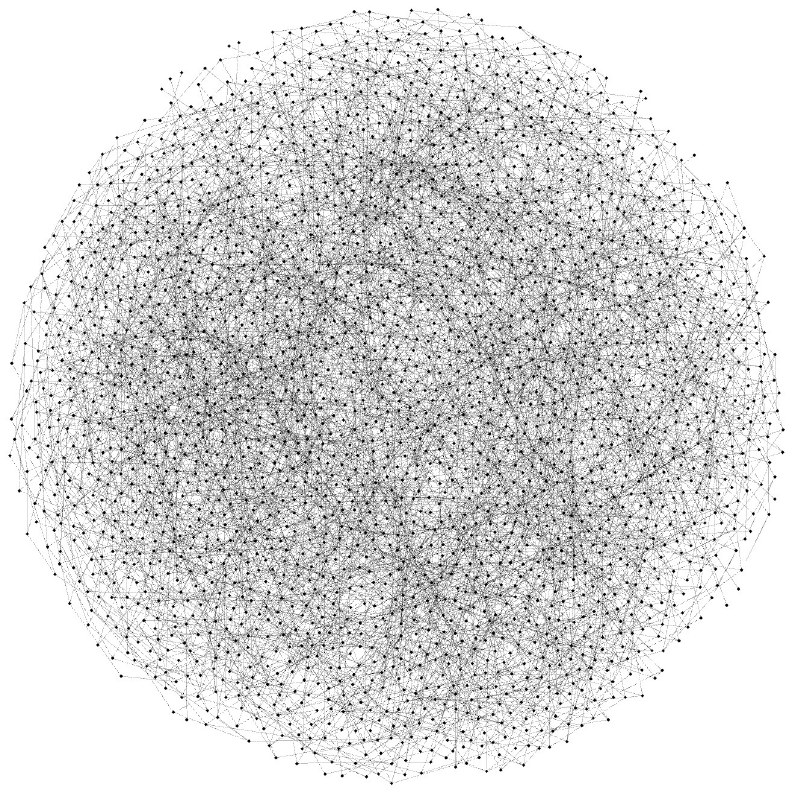}}
    \subfigure[G2SAT] {\includegraphics[width=0.19\linewidth]{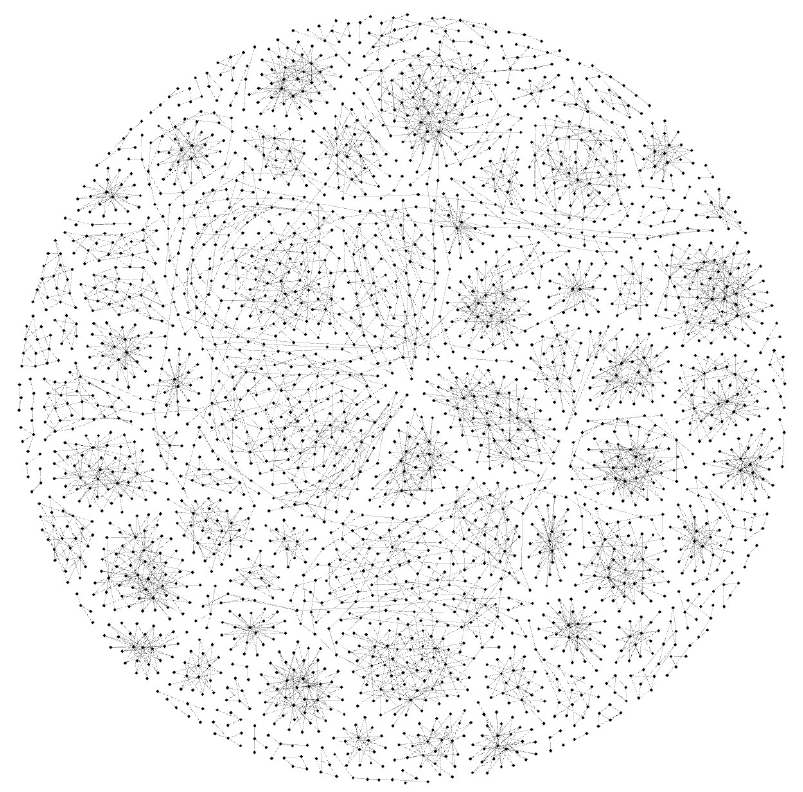}}
    \subfigure[EGNN2S] {\includegraphics[width=0.19\linewidth]{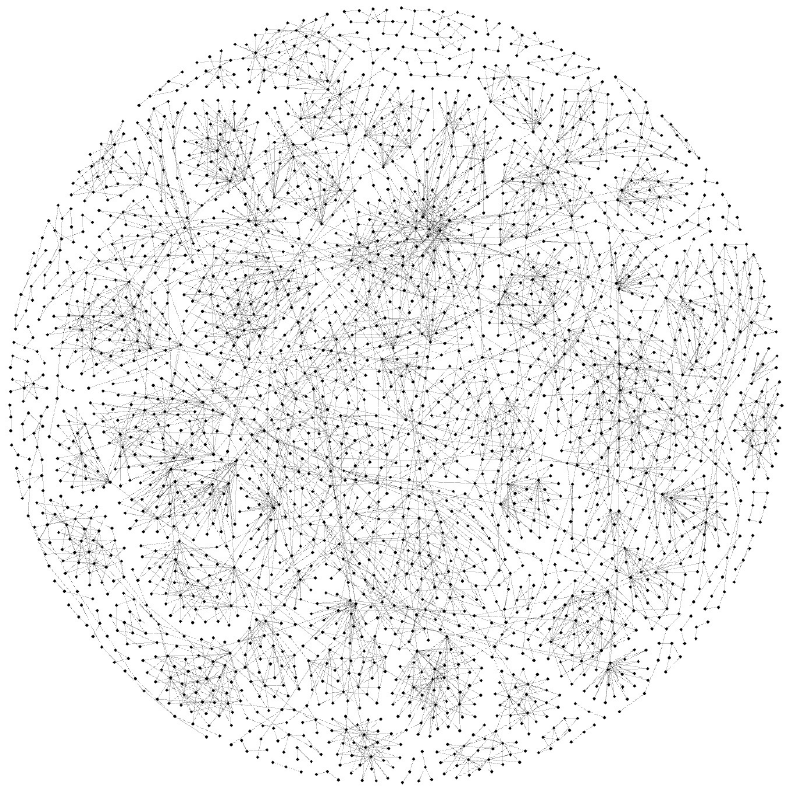}}
    \subfigure[{\mymodel} (SAGE)] {\includegraphics[width=0.19\linewidth]{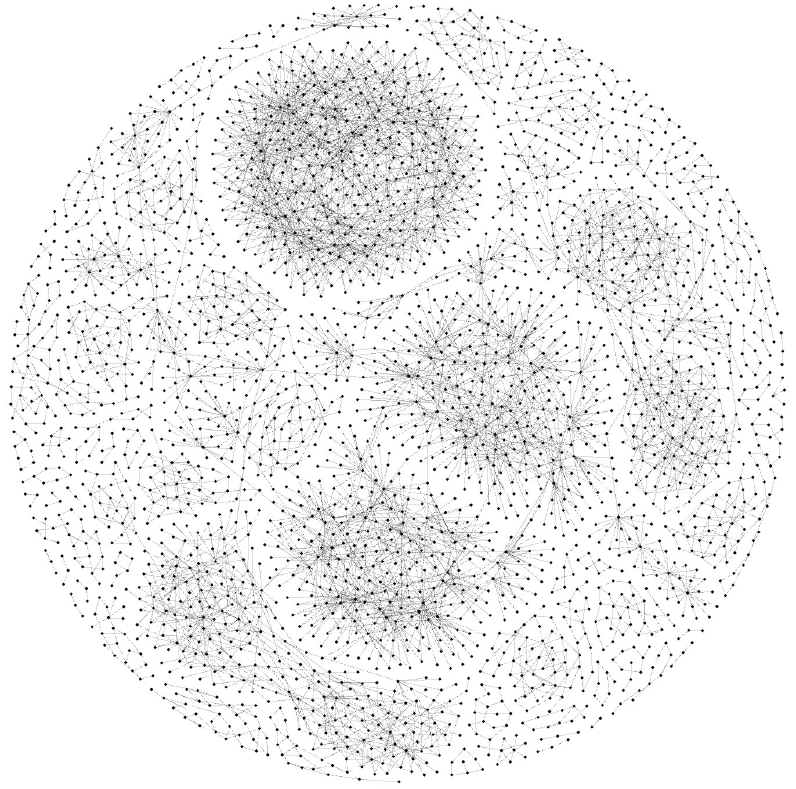}}
    \subfigure[GT] {\includegraphics[width=0.19\linewidth]{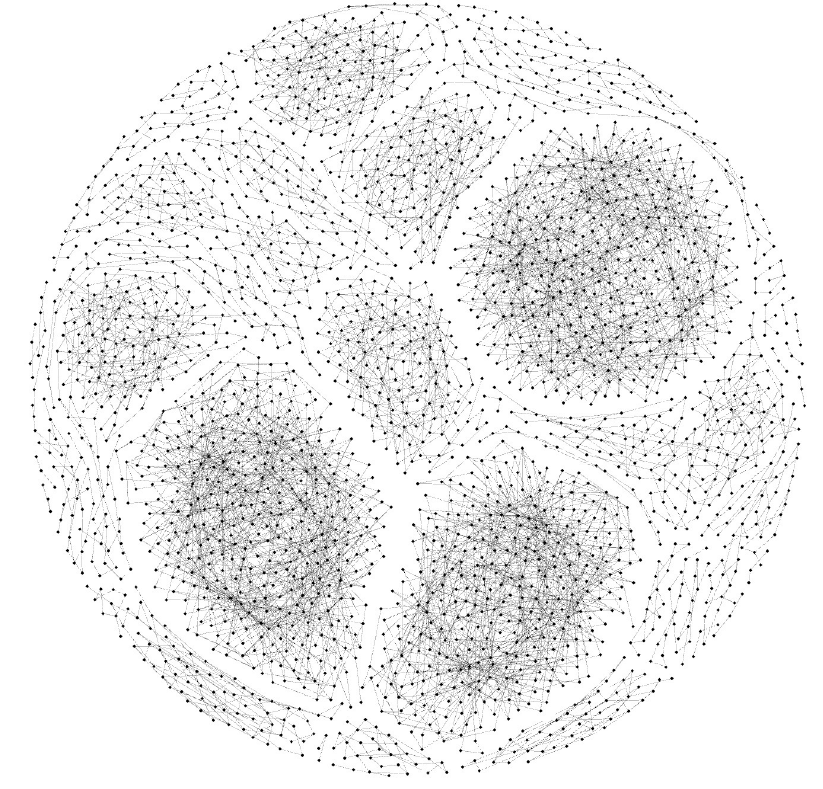}}
    \vspace{-10pt}
    \caption{Visualization of the LCG representations of the generated formulae (w/o post-processing).}
    \label{fig:graphs}
\end{figure*}

\begin{table*}[tb!]
    \centering
    \caption{Results of Kissat performance measured by CPU runtime. Best match in bold.}
    \vspace{-10pt}
    \resizebox{1\linewidth}{!}{
\begin{tabular}{c|cccccc|cc|c}
    \toprule

Instance  & CA~\cite{giraldez2015modularity} & PS~\cite{giraldez2017locality} & G2SAT~\cite{you2019g2sat} & GCN2S~\cite{garzon2022performance} & EGNN2S~\cite{garzon2022performance} & ECC2S~\cite{garzon2022performance} & {\mymodel} (GCN) & {\mymodel} (SAGE) & Ground Truth\\\midrule

lec-1  & 0.45$\pm$0.29 (98.82\%) & 0.48$\pm$0.16 (98.73\%) & 0.59$\pm$0.28 (98.44\%) & 0.47$\pm$0.23 (98.76\%) & 0.51$\pm$0.07 (98.66\%) & 0.47$\pm$0.26 (98.77\%) & \textbf{37.59$\pm$1.84 (1.02\%)} & 40.65$\pm$3.57  (7.03\%) & 37.98  \\

lec-2 & 0.70$\pm$0.28 (95.07\%) & 0.50$\pm$0.28 (96.47\%) & 0.48$\pm$0.24 (96.65\%) & 0.50$\pm$0.28 (96.51\%) & 0.64$\pm$0.20 (95.47\%) & 0.48$\pm$0.29 (96.61\%) & 13.39$\pm$0.31 (5.87\%) & \textbf{14.08$\pm$0.54 (1.04\%)} & 14.23 \\

lec-3  & 1.39$\pm$0.50  (98.55\%) & 0.36$\pm$0.31 (99.63\%)  & 0.30$\pm$0.07 (99.68\%) & 0.42$\pm$0.25 (99.56\%) & 0.60$\pm$0.14 (99.37\%) & 0.62$\pm$0.27 (95.54\%) & 93.51$\pm$6.20  (2.12\%) & \textbf{95.08$\pm$1.61 (0.48\%)} & 95.54 \\

lec-4  & 3.61$\pm$2.99 (92.21\%) & 0.65$\pm$0.29 (98.61\%) & 0.59$\pm$0.25 (98.72\%) & 0.64$\pm$0.29 (98.62\%) & 0.30$\pm$0.17 (99.35\%) & 0.39$\pm$0.19 (99.15\%) & \textbf{44.45$\pm$1.58 (4.13\%)} & 43.22$\pm$1.74  (6.79\%) & 46.37 \\

lec-5  & 0.53$\pm$0.34 (99.08\%) & 0.35$\pm$0.29 (99.39\%) & 0.52$\pm$0.28 (99.10\%) & 0.63$\pm$0.30 (98.92\%) & 0.65$\pm$0.23 (98.87\%) & 0.55$\pm$0.27 (99.05\%) & 52.19$\pm$6.30 (9.94\%) & \textbf{58.39$\pm$4.04  (0.75\%)} & 57.95 \\

lec-6  & 0.46$\pm$0.28 (99.85\%) & 0.47$\pm$0.32 (99.85\%) & 0.43$\pm$0.27 (99.86\%) & 0.36$\pm$0.24 (99.89\%) & 0.54$\pm$0.30 (99.83\%) & 0.41$\pm$0.32 (99.87\%) & 292.76$\pm$27.05 (6.85\%) & \textbf{321.43$\pm$17.84 (2.27\%)} & 314.3\\

lec-7  & 29.36$\pm$31.37 (7.69\%) & 0.65$\pm$0.28 (97.97\%) & 0.16$\pm$0.12 (99.50\%) & 0.53$\pm$0.30 (98.32\%) & 0.41$\pm$0.31 (98.72\%) & 0.60$\pm$0.26 (98.10\%) & 30.65$\pm$0.96 (3.65\%) & \textbf{31.35$\pm$0.87 (1.45\%)} & 31.81 \\

lec-8  & 47.20$\pm$32.63 (59.54\%) & 0.61$\pm$0.29 (99.47\%) & 0.47$\pm$0.27 (98.60\%) & 0.59$\pm$0.26 (99.50\%) & 0.48$\pm$0.40 (99.59\%) & 0.61$\pm$0.25 (99.47\%) & 104.99$\pm$6.44 (10.01\%) & \textbf{112.57$\pm$4.69 (3.50\%)} & 116.66 \\

    \bottomrule
		\end{tabular}
 	}
    \label{tab:hardness_kissat}
\end{table*}

\begin{table*}[tb!]
    \centering
    \caption{Results of Glucose performance measured by CPU runtime with a timeout of 1200 seconds. Best match in bold.}
    \vspace{-10pt}
    \resizebox{1\linewidth}{!}{
\begin{tabular}{c|cccccc|cc|c}
    \toprule

Instance  & CA~\cite{giraldez2015modularity} & PS~\cite{giraldez2017locality} & G2SAT~\cite{you2019g2sat} & GCN2S~\cite{garzon2022performance} & EGNN2S~\cite{garzon2022performance} & ECC2S~\cite{garzon2022performance} & {\mymodel} (GCN) & {\mymodel} (SAGE) & Ground Truth\\\midrule

lec-1  & 0.73$\pm$0.26 (99.81\%) & 0.33$\pm$0.16 (99.91\%) & 0.36$\pm$0.32 (99.91\%) & 0.32$\pm$0.24 (99.92\%) & 0.18$\pm$0.08 (99.95\%) & 0.55$\pm$0.31 (99.86\%) & \textbf{415.19$\pm$31.32 (7.80\%)} & 420.89$\pm$36.38 (9.28\%) & 385.13\\

lec-2 & 1.25$\pm$0.81 (98.99\%) & 0.62$\pm$0.36 (99.49\%) & 0.52$\pm$0.23 (99.58\%) & 0.54$\pm$0.25 (99.56\%) & 0.29$\pm$0.19 (99.77\%) & 0.54$\pm$0.27 (99.56\%) & 105.03$\pm$3.00 (14.73\%) & \textbf{108.12$\pm$2.56 (12.22\%)} & 123.17\\

lec-3  & 40.79$\pm$41.52 (94.66\%) & 0.64$\pm$0.31 (99.92\%)  & 0.72$\pm$0.35 (99.91\%) & 0.46$\pm$0.26 (99.94\%) & 0.29$\pm$0.14 (99.96\%) & 0.70$\pm$0.29 (99.91\%) & 818.75$\pm$30.56 (7.16\%) & \textbf{816.89$\pm$20.75  (6.92\%)} & 764.03\\

lec-4  & 744.81$\pm$490.05 (116.97\%) & 0.69$\pm$0.20 (99.80\%) & 0.28$\pm$0.17 (99.92\%) & 0.50$\pm$0.29 (99.86\%) & 0.57$\pm$0.35 (99.84\%) & 0.43$\pm$0.19 (99.87\%) & 331.76$\pm$14.14 (3.36\%) & \textbf{350.60$\pm$23.02 (2.13\%)} & 343.28\\

lec-5  & 0.57$\pm$0.14 (99.91\%) & 0.27$\pm$0.34 (99.96\%) & 0.54$\pm$0.32 (99.92\%) & 0.68$\pm$0.26 (99.90\%) & 0.51$\pm$0.21 (99.92\%) & 0.43$\pm$0.27 (99.93\%) & 765.95$\pm$131.18 (16.81\%) & \textbf{751.10$\pm$99.93  (14.55\%)} & 655.70\\

lec-6  & 0.90$\pm$0.59 (99.92\%) & 0.35$\pm$0.29 (99.97\%) & 0.61$\pm$0.32 (99.95\%) & 0.62$\pm$0.26 (99.95\%) & 0.62$\pm$0.25 (99.95\%) & 0.50$\pm$0.35 (99.96\%) & \textbf{1200.00 $\pm$0.00 (0.00\%)} & \textbf{1200.00 $\pm$0.00 (0.00\%)} & 1200.00 \\

lec-7  & 1200.00$\pm$0.00 (309.08\%) & 0.50$\pm$0.27 (99.83\%) & 0.68$\pm$0.12 (99.77\%) & 0.60$\pm$0.23 (99.80\%) & 0.69$\pm$0.23 (99.77\%) & 0.68$\pm$0.29 (99.77\%) & \textbf{299.10$\pm$16.38  (1.96\%)} & 307.98$\pm$13.05 (4.99\%) & 293.34 \\

lec-8  & 1200.00$\pm$0.00 (15.52\%) & 0.75$\pm$0.23 (99.93\%) & 0.43$\pm$0.26 (99.96\%) & 0.46$\pm$0.26 (99.96\%) & 0.78$\pm$0.09 (99.92\%) & 0.69$\pm$0.26 (99.93\%) & \textbf{1137.64$\pm$66.58  (9.52\%)} & 1160.04$\pm$30.21 (11.67\%) & 1038.77 \\

    \bottomrule
		\end{tabular}
 	}
    \label{tab:hardness_glucose}
\end{table*}

\begin{table*}[tb!]
    \centering
    \caption{Relative SAT solver performance on ground truth as well as synthetic  formulae. Correct match in underline.}
    \vspace{-10pt}
    \resizebox{0.75\linewidth}{!}{
\begin{tabular}{c|ccccccc|c}
    \toprule

Instance  & CA~\cite{giraldez2015modularity} & PS~\cite{giraldez2017locality} & G2SAT~\cite{you2019g2sat} & GCN2S~\cite{garzon2022performance} & EGNN2S~\cite{garzon2022performance} & ECC2S~\cite{garzon2022performance} & {\mymodel}  & Ground Truth\\\midrule

lec-1  & \underline{($S_1$, $S_2$, $S_3$)} & ($S_3$, $S_1$, $S_2$) & ($S_2$, $S_3$, $S_1$) & ($S_2$, $S_3$, $S_1$) & ($S_3$, $S_1$, $S_2$) & ($S_1$, $S_3$, $S_2$) & \underline{($S_1$, $S_2$, $S_3$)} & ($S_1$, $S_2$, $S_3$) \\

lec-2 & ($S_2$, $S_1$, $S_3$) & ($S_2$, $S_1$, $S_3$) & ($S_1$, $S_3$, $S_2$) & \underline{($S_1$, $S_2$, $S_3$)} & ($S_3$, $S_2$, $S_1$) & ($S_1$, $S_3$, $S_2$) & \underline{($S_1$, $S_2$, $S_3$)} & ($S_1$, $S_2$, $S_3$)\\

lec-3  & \underline{($S_1$, $S_2$, $S_3$)} & \underline{($S_1$, $S_2$, $S_3$)} & ($S_1$, $S_3$, $S_2$) & ($S_2$, $S_1$, $S_3$) & ($S_3$, $S_2$, $S_1$) & \underline{($S_1$, $S_2$, $S_3$)} & \underline{($S_1$, $S_2$, $S_3$)} & ($S_1$, $S_2$, $S_3$)\\

lec-4  & ($S_2$, $S_1$, $S_3$) & ($S_2$, $S_1$, $S_3$) & ($S_3$, $S_2$, $S_1$) & ($S_2$, $S_3$, $S_1$) & \underline{($S_1$, $S_2$, $S_3$)} & ($S_1$, $S_3$, $S_2$) & \underline{($S_1$, $S_2$, $S_3$)} & ($S_1$, $S_2$, $S_3$)\\

lec-5  & ($S_1$, $S_3$, $S_2$) & ($S_3$, $S_1$, $S_2$) & ($S_1$, $S_3$, $S_2$) & \underline{($S_1$, $S_2$, $S_3$)} & ($S_2$, $S_3$, $S_1$) & ($S_2$, $S_3$, $S_1$) & \underline{($S_1$, $S_2$, $S_3$)} & ($S_1$, $S_2$, $S_3$)\\

lec-6  & ($S_1$, $S_2$, $S_3$) & ($S_3$, $S_1$, $S_2$) & ($S_1$, $S_2$, $S_3$) & ($S_1$, $S_2$, $S_3$) & \underline{($S_2$, $S_1$, $S_3$)} & ($S_1$, $S_3$, $S_2$) & \underline{($S_2$, $S_1$, $S_3$)} & ($S_1$, $S_2$, $S_3$)\\

lec-7  & \underline{($S_1$, $S_2$, $S_3$)} & ($S_2$, $S_3$, $S_1$) & \underline{($S_1$, $S_2$, $S_3$)} & \underline{($S_1$, $S_2$, $S_3$)} & \underline{($S_1$, $S_2$, $S_3$)} & ($S_2$, $S_1$, $S_3$) & \underline{($S_1$, $S_2$, $S_3$)} & ($S_1$, $S_2$, $S_3$)\\

lec-8  & ($S_1$, $S_2$, $S_3$) & \underline{($S_2$, $S_1$, $S_3$)} & \underline{($S_2$, $S_1$, $S_3$)} & ($S_3$, $S_2$, $S_1$) & ($S_1$, $S_2$, $S_3$) & \underline{($S_2$, $S_1$, $S_3$)} & \underline{($S_1$, $S_2$, $S_3$)} & ($S_1$, $S_2$, $S_3$) \\

    \bottomrule
		\end{tabular}
 	}
    \label{tab:rank2}
\end{table*}

Fig.~\ref{fig:curves} shows the learning curves of {\mymodel} with SAGE backbone on MP1 datasets. The presented results include the loss, classification AUC of training data, classification ACC of training data, AUC of testing data, and ACC of testing data.

\subsection{Visualization of the Generations}\label{app:vis}
To directly visualize the generated formulae, we visualize the graph structures of the LCG representations via the open-source network analysis and visualization tool Gephi (\url{https://gephi.org/}). Fig.~\ref{fig:graphs} shows the generated graphs of {\mymodel} and baselines on top of the same graph template. The graphs are laid out by Fruchterman–Reingold graph drawing algorithm~\cite{fruchterman1991graph}.

\subsection{More Results of Solver Runtime}\label{app:more_runtime}

Table ~\ref{tab:hardness_kissat} and Table ~\ref{tab:hardness_glucose} show the evaluation of Kissat~\cite{fleury2020cadical} and Glucose~\cite{audemard2009glucose} runtime over the generated formulae as well as the original benchmarks. The experimental settings are the same as Table ~\ref{tab:hardness}. The three tables are the supporting evidence to derive Table ~\ref{tab:rank}. As seen, {\mymodel} recover the computational hardness of the generated formulae with respect to all three mainstream SAT solvers.\looseness=-1

\subsection{More Results of Solver Ranking}\label{app:more_rank}

Table ~\ref{tab:rank2} shows the relative SAT solver performance on ground truth formulae and the generated formulae. The same solver ranking to the ground truth indicates better recovery of the computational properties of the SAT formulae. The experimental setting are the same as Table ~\ref{tab:rank}. As seen, {\mymodel} is the only method to recover the rankings on all templates. 


\end{document}